\documentclass[lettersize,journal]{IEEEtran}
\usepackage{amsmath,amsfonts}
\usepackage{amssymb}
\usepackage{amsthm}   
\usepackage{algorithm}
\usepackage[noend]{algpseudocode}
\usepackage{array}
\usepackage[caption=false,font=normalsize,labelfont=sf,textfont=sf]{subfig}
\usepackage{textcomp}
\usepackage{pifont}
\usepackage{cleveref}
\usepackage{stfloats}
\usepackage{url}
\usepackage{verbatim}
\usepackage{graphicx}
\usepackage{cite}
\usepackage{marvosym}
\usepackage{tabularx}
\usepackage{makecell}
\usepackage{colortbl}
\usepackage{booktabs}
\usepackage{multirow}
\usepackage[normalem]{ulem}
\usepackage[caption=false]{subfig}
\usepackage{xcolor}
\usepackage{float}
\usepackage{rotating}

\definecolor{ManBlue}{RGB}{222,240,255}

\newcolumntype{M}{>{\columncolor{ManBlue}}c}

\usepackage[normalem]{ulem}
\usepackage{cancel}

\DeclareRobustCommand{\DEL}[1]{{\color{red}\ifmmode\cancel{#1}\else\sout{#1}\fi}}

\NewDocumentCommand{\tyh}
{ mO{} }{\textcolor{blue}{\textsuperscript{\textit{tyh}}\textsf{\textbf{\small[#1]}}}}

\begin{document}

\title{Replacement Learning: Training Neural Networks with Fewer Parameters}

\author{Yuming Zhang, Peizhe Wang, Tianyang Han, Hengyu Shi, Junhao Su, Dongzhi Guan, Jiabin Liu, Jiaji Wang

\thanks{Yuming Zhang is a PhD student in The University of Hong Kong, Hong Kong, China. Email: yuming\_zhang@connect.hku.hk}
\thanks{Tianyang Han, Hengyu Shi, and Junhao Su are now independent researchers. Email: \{hantianyang522, qq1842084, junh.suuu\}@gmail.com}
\thanks{Peizhe Wang is a Master's student in Southeast University, Nanjing, China. Email: 220221387@seu.edn.cn}
\thanks{Dongzhi Guan and Jiabin Liu are Associate Professors in Southeast University, Nanjing, China. Email: guandongzhi@seu.edu.cn, dndxljb@126.com}
\thanks{Jiaji Wang is an Assistant Professor in The University of Hong Kong, Hong Kong, China. Email: cewang@hku.hk}

\thanks{Yuming Zhang, Peizhe Wang, Hengyu Shi, Tianyang Han did this work with equal contributions. }
\thanks{The project leader is Junhao Su.}
\thanks{The corresponding author is Jiaji Wang.}
\thanks{The work in this paper was financially supported by the National Natural Science Foundation of China (Grant No. 52408221), Hong Kong Innovation and Technology Support Programme (Mid-stream, theme-based, ITS/041/23MX), and Hong Kong Environment and Conservation Fund (ECF Project 249/2024).}
}

\markboth{IEEE}
{Shell \MakeLowercase{\textit{et al.}}: A Sample Article Using IEEEtran.cls for IEEE Journals}


\maketitle

\begin{abstract}
End-to-end training with full-depth backpropagation remains the dominant paradigm for optimizing deep neural networks, but its efficiency deteriorates as models grow deeper. Since every block must be executed and differentiated under a single global objective, full-depth BP introduces substantial parameter redundancy, activation-memory cost, and training latency, especially when neighboring layers exhibit highly correlated learning patterns. Directly skipping or removing layers can reduce cost, but often weakens representation capacity or requires architecture-specific reuse designs. In this paper, we propose \textbf{Replacement Learning} (RepL), a training-time paradigm that reduces full-depth redundancy by replacing selected blocks rather than simply discarding them. For each removed block, RepL inserts a lightweight computing layer that synthesizes a surrogate operator from the parameters of its adjacent preceding and succeeding blocks through a learnable transformation, and applies the synthesized operator to the preceding activation. In this way, RepL preserves local contextual continuity while avoiding unnecessary full-layer computation. We instantiate RepL for CNNs and ViTs with tailored parameter-fusion blocks that handle convolutional channels, feature resolutions, and transformer submodules. Extensive experiments on CIFAR-10, SVHN, STL-10, ImageNet, COCO, and CityScapes show that RepL reduces trainable parameters, GPU memory usage, and training time while matching or surpassing standard end-to-end training across classification, detection, and segmentation. Additional results on WikiText-2, transfer learning, inference throughput, checkpointing, stochastic depth, and INT8 quantization further demonstrate its generality and compatibility.
\end{abstract}

\section{Introduction}
\label{sec:intro}
\begin{figure*}[h]
  \centering
  \includegraphics[width=0.99\textwidth]{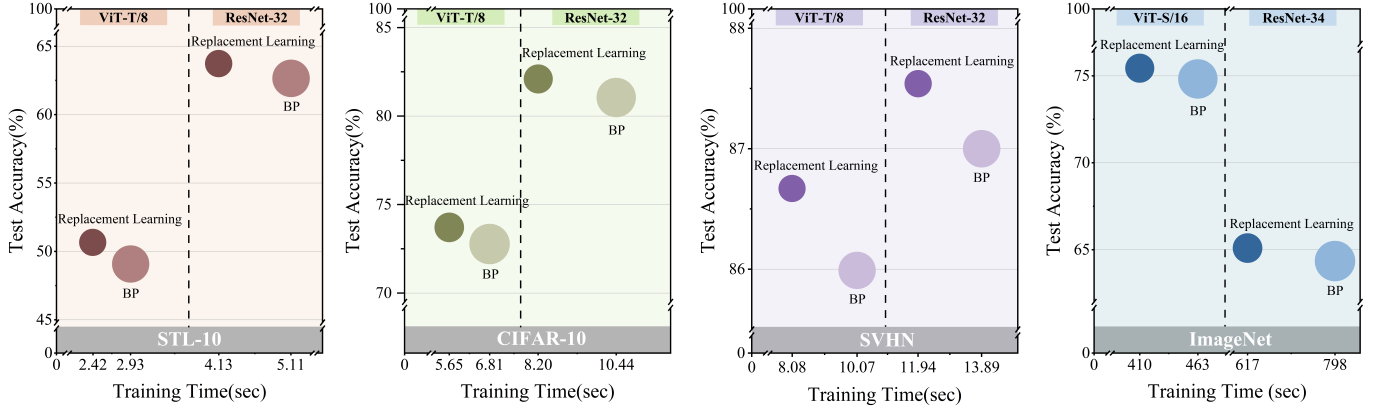}
  \caption{Comparison between different backbones with Replacement Learning and End-to-End training regarding GPU memory and Test accuracy. The diameter of the symbol is obtained based on GPU Memory at the same scale.}
  \label{p1}
\end{figure*}
Updating learnable parameters is fundamental for training deep learning models \cite{r35}. The most common method, global backpropagation \cite{r28}, is widely applied in fields like computer vision \cite{r29, r30}, natural language processing \cite{r31, r32}, and speech processing \cite{r33, r34}. However, increasing model capabilities inevitably raises network depth and complexity, sharply escalating the computational and parameter demands of global backpropagation \cite{r39}, which challenges GPU processing power and memory capacity \cite{r37}. Moreover, high similarity in learning patterns between neighbouring layers \cite{r40} causes parameter redundancy and inefficient resource usage. With large models becoming prevalent, developing effective training methods to reduce computation time and save GPU memory while preserving performance is urgently needed.
 
To tackle the challenges of traditional backpropagation (BP) \cite{r28}, researchers have explored alternatives such as feedback alignment \cite{r22, r23}, forward gradient learning \cite{r24, r25}, and local learning \cite{r26, r27,guo2024faster}. These methods aim to update network weights without fully relying on BP \cite{r12}, thereby reducing training costs. However, they each have limitations. Feedback alignment struggles with training effectiveness due to inaccurate gradient estimation. Forward gradient learning requires extra forward passes, increasing computational overhead. Local learning divides the network into independently trained modules, but this often leads to suboptimal local performance and longer training times. Recent work on Vision Transformers (ViTs) \cite{r14} revealed strong inter-layer correlations from self-attention, leading to the skip attention \cite{r21} approach to reduce complexity by reusing attention computations. However, this method requires manually designed auxiliary modules, making it complex and hard to generalize. Additionally, it risks error propagation, negatively impacting model performance. As a result, alternatives to backpropagation \cite{r12} and skip attention \cite{r21} still face challenges in balancing training efficiency and computational cost while maintaining performance.

\begin{figure*}[h]
  \centering
\includegraphics[width=0.989\textwidth] {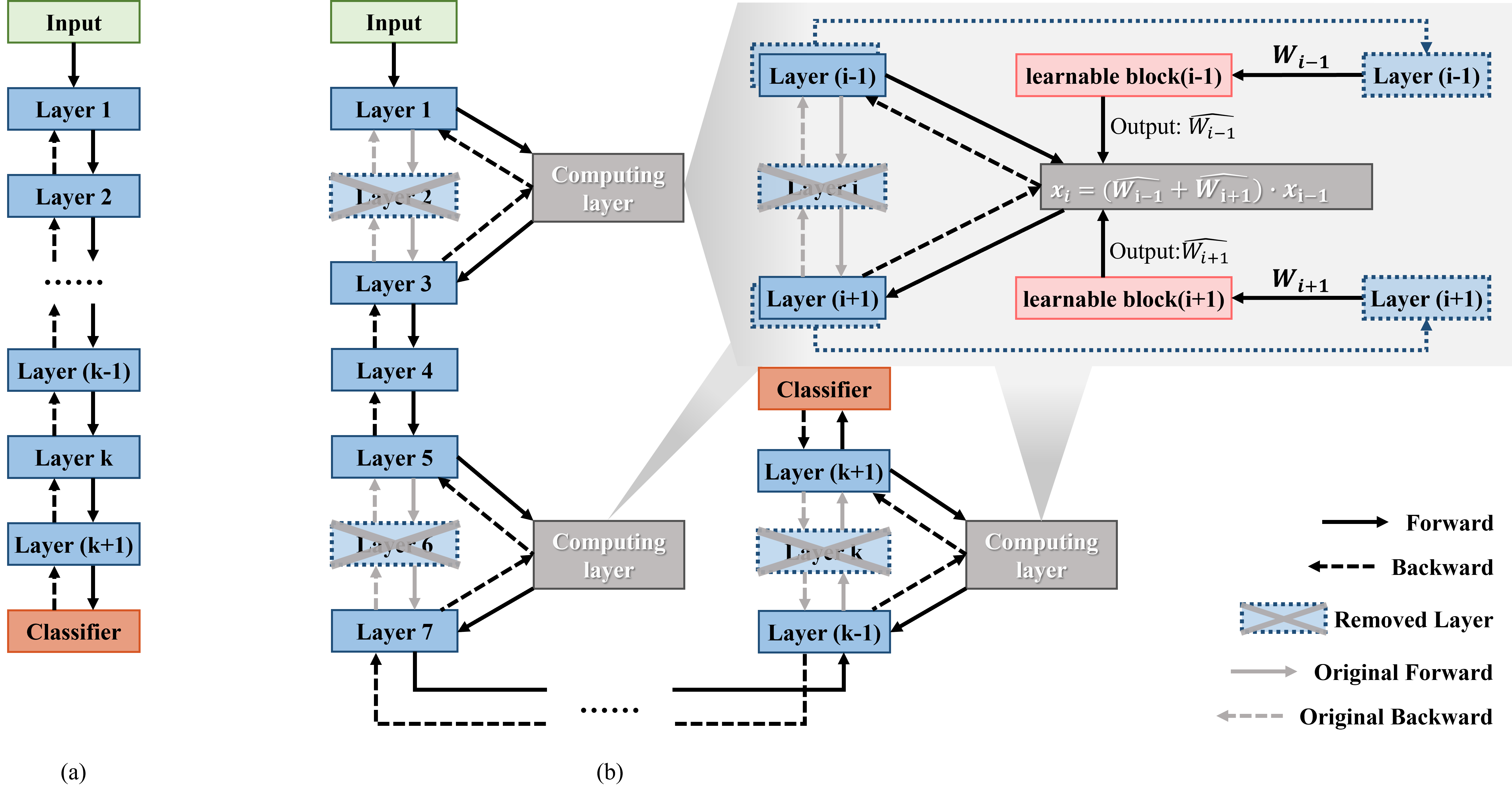}
  \caption{Comparison of (a) End-to-End training and (b) our proposed Replacement Learning.}
  \label{p2}
\end{figure*}

In this paper, we propose a novel method: Replacement Learning (RepL), which aims to reduce the computational overhead and resource consumption of deep neural networks while maintaining—or even improving—model performance. The framework is shown in \cref{p2}. The core idea of RepL is to selectively remove specific layers of the network and replace them with a lightweight computing layer that features a simple structure and minimal parameter count. Specifically, the computing layer synthesizes new computational parameters by integrating information from the parameters of the layers immediately preceding and succeeding the removed layer. This integration is accomplished through a specially designed, lightweight, learnable block. The fused parameters are then used to reprocess the output of the preceding layer, which is subsequently fed into the succeeding layer. The design notably enhances the network’s capacity to capture local features in shallow layers and global representations in deeper layers, thereby promoting a more effective integration of low-level and high-level features. Moreover, we introduce an optimized interval strategy to regulate the frequency at which layers are removed and optimized, striking a desirable balance between computational efficiency and model performance. By leveraging two specially designed learnable blocks within the computing layer, RepL achieves efficient fusion of adjacent layer information and dynamically balances the retention of historical context with the incorporation of new feature representations, thereby further boosting overall performance.
We comprehensively evaluate the effectiveness of RepL on five widely used benchmark datasets-CIFAR-10 \cite{r1}, STL-10 \cite{r3}, SVHN \cite{r2}, ImageNet \cite{r4}, and COCO \cite{r51}—across image classification and object detection tasks, employing both CNNs and ViTs \cite{r14} architectures. Experimental results demonstrate that, compared with traditional End-to-End training methods \cite{r12}, RepL not only significantly reduces the number of trainable parameters, training time, and GPU memory usage, but also achieves superior performance in terms of model accuracy.
We summarize our contributions as follows:
\begin{itemize}
\item We propose a novel and general training method, RepL, which achieves performance comparable to or even surpassing that of traditional End-to-End training methods \cite{r12}, while significantly reducing the number of parameters, training time, and GPU memory consumption.
\item RepL is applicable to both CNNs and ViTs and extensible across multiple tasks, exhibiting strong generalizability. It can be flexibly applied to models of varying depths and across different domains.
\item We conduct extensive experiments on several widely-used image classification and object detection benchmarks, including CIFAR-10 \cite{r1}, STL-10 \cite{r3}, SVHN \cite{r2}, ImageNet \cite{r4}, and COCO \cite{r51}. Results demonstrate that Replacement Learning consistently outperforms traditional End-to-End training methods in both computational efficiency and model performance.
\end{itemize}
\section{Related Work}
\label{sec:formatting}
\textbf{Scope and non-goals.}
Our work proposes a training-time modification intended as a drop-in alternative to standard end-to-end (E2E) backpropagation for a \textbf{fixed} backbone. We do \textbf{not} aim to produce a deployment-time compressor, nor to compete with post-hoc model compression or search pipelines such as pruning \cite{depgraph}, distillation \cite{distilling}, quantization \cite{quantization}, or NAS \cite{darts}.
These methods are orthogonal to the objective of this work and can be further applied on top of models trained using our approach; we provide experimental evidence for this claim in the supplementary material \ref{sec:repl_quantization}. In summary, this paper focuses on \textbf{training strategies}, rather than deployment-time inference.

\subsection{Alternatives to backpropagation} 
To address the inherent limitations of backpropagation, such as high computational cost, various alternative methods have been proposed, including target propagation \cite{r47, r48}, feedback alignment \cite{r22, r23}, and decoupled neural interfaces (DNI) \cite{r49}. These approaches bypass traditional global backpropagation by directly propagating errors to individual layers, reducing memory usage and enhancing efficiency. Forward gradient learning \cite{r24, r25} offers a new paradigm for training deep networks more effectively. Local learning \cite{r41, su2025man++,glcan} segments the network into smaller, independently trained modules, optimizing local objectives to lower computational demands while preserving some global features \cite{r26, r27,lopt,trace}. However, excessive segmentation may give rise to inter-module coordination challenges. This misalignment ultimately undermines the model’s overall performance, particularly when dealing with complex datasets like ImageNet that demand robust cross-module feature integration.

\subsection{Utilizing surrounding layers} 
Leveraging the strikingly high similarity in learning conditions of surrounding layers, researchers have solved many problems in deep learning. Some studies have applied  Residual Networks (ResNets) \cite{r5}, by adding a shortcut connection to the activation function of the next layer, this identity mapping enables ResNet to effectively mitigate the notorious problem of performance degradation \cite{r43, r44}, enhancing both the convergence speed and accuracy of the network \cite{r45, r46}. Additionally, some researchers have proposed skipping attention, reusing the self-attention calculations from one layer in the approximations for attention in subsequent layers, which in turn translates into substantially higher computational throughput for the model \cite{r21}. However, due to the repeated use of prior layers, this method carries the risk of error propagation and could potentially cause losses during the learning process, impacting the model's generalization ability.

\section{Method}
\label{sec:method}

We propose \textbf{Replacement Learning (RepL)}, a structured
training-time replacement paradigm for deep neural networks. RepL
physically removes a subset of internal blocks from the original
backbone and inserts a lightweight computing layer at each removed
position. The inserted computing layer is not an independently learned
full block. Instead, it synthesizes its transformation from the
parameters of the adjacent retained blocks. In this way, RepL reduces
the redundancy of full-depth training while preserving the local
continuity of the original network.

Throughout this section, we use $K$ to denote the replacement interval
and use $q$ to denote the spatial convolution kernel size. In our main
experiments, we set $K=4$.

\subsection{Replacement Learning Framework}
\label{sec:framework}

Let a network of depth $L$ be written as
\begin{equation}
\begin{aligned}
  F(\mathbf{x})
  &=
  f_L \circ f_{L-1}
  \circ \cdots \circ f_1(\mathbf{x}),       \\
  \mathbf{h}_{\ell}
  &=
  f_{\ell}
  \!\left(
    \mathbf{h}_{\ell-1};
    \theta_{\ell}
  \right),
  \qquad
  \mathbf{h}_0=\mathbf{x}.
\end{aligned}
\label{eq:standard_forward}
\end{equation}
Here $f_{\ell}$ denotes either a convolutional residual block or a
transformer block, and $\theta_{\ell}$ denotes its learnable
parameters.

RepL removes one internal block every $K$ blocks. The removed block set
is defined as
\begin{equation}
\begin{aligned}
  \mathcal{R}
  =
  \left\{
  r
  \ \middle|\
  r=mK,\;
  1 \le r < L,\;
  m=1,2,\ldots
  \right\}.
\end{aligned}
\label{eq:removed_set}
\end{equation}
The last block is kept even if it satisfies the interval condition,
because the computing layer at position $r$ requires both a left
neighbor and a right neighbor.

For each $r\in\mathcal{R}$, the original block $f_r$ is physically
removed. The local computation
\begin{equation}
\begin{aligned}
  f_{r+1}
  \circ
  f_r
  \circ
  f_{r-1}
\end{aligned}
\end{equation}
is replaced by
\begin{equation}
\begin{aligned}
  f_{r+1}
  \circ
  g_r
  \circ
  f_{r-1},
\end{aligned}
\label{eq:local_replacement}
\end{equation}
where $g_r$ is a lightweight computing layer inserted between the two
adjacent retained blocks.

The parameters of $g_r$ are synthesized from the neighboring block
parameters:
\begin{equation}
\begin{aligned}
  \widehat{\theta}_r
  =
  \mathcal{T}^{\mathrm{prev}}_r
  \!\left(
    \theta_{r-1}
  \right)
  +
  \mathcal{T}^{\mathrm{next}}_r
  \!\left(
    \theta_{r+1}
  \right).
\end{aligned}
\label{eq:general_synthesis}
\end{equation}
Here $\mathcal{T}^{\mathrm{prev}}_r$ and
$\mathcal{T}^{\mathrm{next}}_r$ are lightweight learnable
transformations in weight space. The computing layer then applies the
synthesized operator to the current activation:
\begin{equation}
\begin{aligned}
  g_r(\mathbf{h})
  =
  \Phi_r
  \!\left(
    \mathbf{h};
    \widehat{\theta}_r
  \right),
\end{aligned}
\label{eq:general_computing_layer}
\end{equation}
where $\Phi_r$ is instantiated differently for CNNs and ViTs.

\subsection{RepL for CNNs}
\label{sec:cnn_repl}

We first instantiate RepL for ResNet-style CNNs. To avoid incompatible
feature shapes, replacement is performed only within the same residual
stage. Stage-transition blocks and the last block of each stage are
kept, so that the left and right neighbors of each removed block have
compatible spatial resolutions and channel dimensions.

\subsubsection{BasicBlock Replacement}
\label{sec:basicblock_repl}

A standard ResNet BasicBlock can be written as
\begin{equation}
\begin{aligned}
  \mathbf{z}_{\ell}
  &=
  \rho
  \!\left(
    \operatorname{BN}_{\ell,1}
    \!\left(
      W_{\ell,1}^{3\times3}
      *
      \mathbf{x}_{\ell}
    \right)
  \right),                                               \\
  \mathbf{u}_{\ell}
  &=
  \operatorname{BN}_{\ell,2}
  \!\left(
    W_{\ell,2}^{3\times3}
    *
    \mathbf{z}_{\ell}
  \right),                                               \\
  f_{\ell}(\mathbf{x}_{\ell})
  &=
  \rho
  \!\left(
    \mathbf{x}_{\ell}
    +
    \mathbf{u}_{\ell}
  \right),
\end{aligned}
\label{eq:basicblock}
\end{equation}
where $*$ denotes convolution and $\rho(\cdot)$ denotes ReLU.

When the $r$-th BasicBlock is removed, RepL uses the second convolution
of the previous block and the first convolution of the next block:
\begin{equation}
\begin{aligned}
  W_{\mathrm{prev}}
  &=
  W_{r-1,2}^{3\times3},                                      \\
  W_{\mathrm{next}}
  &=
  W_{r+1,1}^{3\times3}.
\end{aligned}
\label{eq:basic_neighbor_weights}
\end{equation}

Before fusion, each neighboring kernel is normalized per output
channel. For
$W\in\mathbb{R}^{C_{\mathrm{out}}\times C_{\mathrm{in}}\times q\times q}$,
we define
\begin{equation}
\begin{aligned}
  \overline{W}_{c,:,:,:}
  =
  \frac{
    W_{c,:,:,:}
  }{
    \sqrt{
      \sum_{a,u,v}
      W_{c,a,u,v}^{2}
      +
      \epsilon
    }
  },
  \qquad
  c=1,\ldots,C_{\mathrm{out}} .
\end{aligned}
\label{eq:conv_norm}
\end{equation}

The replacement kernel is synthesized by channel-wise learnable scales:
\begin{equation}
\begin{aligned}
  \widehat{W}_{r,c,:,:}^{3\times3}
  &=
  \alpha_{r,c}
  \overline{W}_{\mathrm{prev},c,:,:}
  +
  \beta_{r,c}
  \overline{W}_{\mathrm{next},c,:,:},                         \\
  c
  &=
  1,\ldots,C_{\mathrm{out}} .
\end{aligned}
\label{eq:basic_kernel_synthesis}
\end{equation}
This formulation avoids a long operator-mapping expression and directly
specifies the channel-wise synthesis used by the computing layer.

The BasicBlock computing layer is
\begin{equation}
\begin{aligned}
  \mathbf{y}_r
  &=
  \operatorname{BN}_{r}
  \!\left(
    \widehat{W}_{r}^{3\times3}
    *
    \mathbf{x}
  \right),                                                  \\
  g_r^{\mathrm{basic}}(\mathbf{x})
  &=
  \rho
  \!\left(
    \mathbf{x}
    +
    \mathbf{y}_r
  \right).
\end{aligned}
\label{eq:basic_forward}
\end{equation}
Thus, compared with the removed BasicBlock, RepL keeps a residual
topology but replaces the full two-convolution block with a single
neighbor-synthesized convolutional computation.

\subsubsection{Bottleneck Replacement}
\label{sec:bottleneck_repl}

For deeper ResNets, a Bottleneck block consists of a $1\times1$
reduction convolution, a $3\times3$ spatial convolution, and a
$1\times1$ expansion convolution:
\begin{equation}
\begin{aligned}
  \mathbf{z}_{\ell,1}
  &=
  \rho
  \!\left(
    \operatorname{BN}_{\ell,1}
    \!\left(
      W_{\ell,1}^{1\times1}
      *
      \mathbf{x}_{\ell}
    \right)
  \right),                                                \\
  \mathbf{z}_{\ell,2}
  &=
  \rho
  \!\left(
    \operatorname{BN}_{\ell,2}
    \!\left(
      W_{\ell,2}^{3\times3}
      *
      \mathbf{z}_{\ell,1}
    \right)
  \right),                                                \\
  \mathbf{u}_{\ell}
  &=
  \operatorname{BN}_{\ell,3}
  \!\left(
    W_{\ell,3}^{1\times1}
    *
    \mathbf{z}_{\ell,2}
  \right),                                                \\
  f_{\ell}(\mathbf{x}_{\ell})
  &=
  \rho
  \!\left(
    \mathbf{x}_{\ell}
    +
    \mathbf{u}_{\ell}
  \right).
\end{aligned}
\label{eq:bottleneck}
\end{equation}

When the $r$-th Bottleneck is removed, RepL uses the adjacent
Bottleneck parameters as follows:
\begin{equation}
\begin{aligned}
  W_{\mathrm{red}}
  &=
  W_{r-1,1}^{1\times1},                                  \\
  W_{\mathrm{mid,prev}}
  &=
  W_{r-1,2}^{3\times3},                                  \\
  W_{\mathrm{mid,next}}
  &=
  W_{r+1,2}^{3\times3},                                  \\
  W_{\mathrm{exp}}
  &=
  W_{r+1,3}^{1\times1}.
\end{aligned}
\label{eq:bottleneck_neighbor_weights}
\end{equation}
The middle spatial kernel is synthesized from the neighboring
$3\times3$ kernels:
\begin{equation}
\begin{aligned}
  \widehat{W}_{r,2,c,:,:}^{3\times3}
  &=
  \alpha_{r,c}
  \overline{W}_{\mathrm{mid,prev},c,:,:}
  +
  \beta_{r,c}
  \overline{W}_{\mathrm{mid,next},c,:,:},                 \\
  c
  &=
  1,\ldots,C_{\mathrm{mid}} .
\end{aligned}
\label{eq:bottleneck_mid_synthesis}
\end{equation}

The Bottleneck computing layer is then
\begin{equation}
\begin{aligned}
  \mathbf{z}_{1}
  &=
  \rho
  \!\left(
    W_{\mathrm{red}}
    *
    \mathbf{x}
  \right),                                                \\
  \mathbf{z}_{2}
  &=
  \rho
  \!\left(
    \widehat{W}_{r,2}^{3\times3}
    *
    \mathbf{z}_{1}
  \right),                                                \\
  \mathbf{y}_r
  &=
  \operatorname{BN}_{r}
  \!\left(
    W_{\mathrm{exp}}
    *
    \mathbf{z}_{2}
  \right),                                                \\
  g_r^{\mathrm{bottle}}(\mathbf{x})
  &=
  \rho
  \!\left(
    \mathbf{x}
    +
    \mathbf{y}_r
  \right).
\end{aligned}
\label{eq:bottleneck_forward}
\end{equation}
If the original Bottleneck uses grouped convolution in the middle
$3\times3$ layer, the synthesized middle convolution follows the same
group setting.

\subsubsection{Stage-wise CNN Execution}
\label{sec:cnn_stage_execution}

For a residual stage with $L_s$ blocks, RepL removes internal blocks at
a fixed interval:
\begin{equation}
\begin{aligned}
  \mathcal{R}_s
  =
  \left\{
  r
  \ \middle|\
  r=mK,\;
  1 \le r < L_s,\;
  m=1,2,\ldots
  \right\}.
\end{aligned}
\label{eq:stage_removed_set}
\end{equation}
After the original blocks in $\mathcal{R}_s$ are physically removed, the forward propagation inside the stage is
\begin{equation}
\begin{aligned}
  \mathbf{x}
  &\leftarrow
  f_{r-1}^{(s)}(\mathbf{x}),                              \\
  \mathbf{x}
  &\leftarrow
  g_{r}^{(s)}
  \!\left(
    \mathbf{x};
    \theta_{r-1}^{(s)},
    \theta_{r+1}^{(s)}
  \right),                                                \\
  \mathbf{x}
  &\leftarrow
  f_{r+1}^{(s)}(\mathbf{x}).
\end{aligned}
\label{eq:cnn_stage_forward}
\end{equation}
No computing layer is constructed across stage boundaries.

\subsection{RepL for Vision Transformers}
\label{sec:vit_repl}

We next instantiate RepL for Vision Transformers. A standard pre-norm ViT block is
\begin{equation}
\begin{aligned}
  \mathbf{U}_{\ell}
  &=
  \mathbf{X}_{\ell}
  +
  \operatorname{MSA}_{\ell}
  \!\left(
    \operatorname{LN}_{\ell,1}
    (\mathbf{X}_{\ell})
  \right),                                                \\
  \mathbf{X}_{\ell+1}
  &=
  \mathbf{U}_{\ell}
  +
  \operatorname{MLP}_{\ell}
  \!\left(
    \operatorname{LN}_{\ell,2}
    (\mathbf{U}_{\ell})
  \right),
\end{aligned}
\label{eq:vit_block}
\end{equation}
where
$\mathbf{X}_{\ell}\in\mathbb{R}^{B\times T\times d}$
contains $T$ tokens with embedding dimension $d$.

For a removed transformer block $r$, RepL mainly uses the attention
output projection matrices from the adjacent retained blocks:
\begin{equation}
\begin{aligned}
  W_{\mathrm{prev}}^{\mathrm{proj}}
  &=
  W_{r-1}^{\mathrm{proj}},                         \\
  W_{\mathrm{next}}^{\mathrm{proj}}
  &=
  W_{r+1}^{\mathrm{proj}},                         \\
  b_{\mathrm{prev}}^{\mathrm{proj}}
  &=
  b_{r-1}^{\mathrm{proj}},                         \\
  b_{\mathrm{next}}^{\mathrm{proj}}
  &=
  b_{r+1}^{\mathrm{proj}} .
\end{aligned}
\label{eq:vit_neighbor_projection}
\end{equation}

\subsubsection{ViT Attention Replacement}
\label{sec:vit_attention_repl}

The synthesized attention projection is
\begin{equation}
\begin{aligned}
  \widehat{W}_{r}^{\mathrm{proj}}
  &=
  \alpha_r
  W_{\mathrm{prev}}^{\mathrm{proj}}
  +
  \beta_r
  W_{\mathrm{next}}^{\mathrm{proj}},                  \\
  \widehat{b}_{r}^{\mathrm{proj}}
  &=
  \alpha_r
  b_{\mathrm{prev}}^{\mathrm{proj}}
  +
  \beta_r
  b_{\mathrm{next}}^{\mathrm{proj}} .
\end{aligned}
\label{eq:vit_projection_synthesis}
\end{equation}
Here $\alpha_r$ and $\beta_r$ are learnable coefficients introduced by
the computing layer.

The ViT computing layer applies a token-wise linear transformation:
\begin{equation}
\begin{aligned}
  \Delta_r
  &=
  \mathbf{X}
  \left(
    \widehat{W}_{r}^{\mathrm{proj}}
  \right)^{\top}
  +
  \widehat{b}_{r}^{\mathrm{proj}},                    \\
  g_r^{\mathrm{vit}}(\mathbf{X})
  &=
  \mathbf{X}
  +
  \Delta_r .
\end{aligned}
\label{eq:vit_basic_forward}
\end{equation}
This replacement keeps the residual form of the transformer block while
removing the full self-attention and MLP computation of the original
block.

\subsubsection{Per-head Weight Synthesis}
\label{sec:vit_head_synthesis}

For larger ViTs, RepL can synthesize the projection matrix in a
head-wise manner. Let the number of heads be $H$, and the head dimension
be $d_h=d/H$. Let $\mathcal{C}_h$ denote the channel index set of the
$h$-th head. We first normalize the neighboring projection matrices:
\begin{equation}
\begin{aligned}
  \overline{W}_{i,:}
  =
  \frac{
    W_{i,:}
  }{
    \sqrt{
      \sum_j W_{i,j}^{2}
      +
      \epsilon
    }
  } .
\end{aligned}
\label{eq:linear_norm}
\end{equation}
The head-wise synthesized projection is
\begin{equation}
\begin{aligned}
  \widehat{W}_{r,:, \mathcal{C}_h}^{\mathrm{proj}}
  &=
  \alpha_{r,h}
  \overline{W}_{\mathrm{prev},:,\mathcal{C}_h}^{\mathrm{proj}}
  \nonumber \\
  &\quad+
  \beta_{r,h}
  \overline{W}_{\mathrm{next},:,\mathcal{C}_h}^{\mathrm{proj}},
  \qquad
  h=1,\ldots,H.
\end{aligned}
\label{eq:vit_headwise_synthesis}
\end{equation}
The corresponding bias is synthesized as
\begin{equation}
\begin{aligned}
  \widehat{b}_{r}^{\mathrm{proj}}
  =
  \frac{1}{2}
  \left(
    b_{\mathrm{prev}}^{\mathrm{proj}}
    +
    b_{\mathrm{next}}^{\mathrm{proj}}
  \right).
\end{aligned}
\label{eq:vit_bias_average}
\end{equation}

With normalization before the replacement branch, the computing layer is
\begin{equation}
\begin{aligned}
  \widetilde{\mathbf{X}}
  &=
  \operatorname{LN}_{r-1,1}
  (\mathbf{X}),                                      \\
  \Delta_r
  &=
  d^{-1/2}
  \widetilde{\mathbf{X}}
  \left(
    \widehat{W}_{r}^{\mathrm{proj}}
  \right)^{\top}
  +
  \widehat{b}_{r}^{\mathrm{proj}},                  \\
  g_r^{\mathrm{vit}}(\mathbf{X})
  &=
  \mathbf{X}
  +
  \Delta_r .
\end{aligned}
\label{eq:vit_head_forward}
\end{equation}

\subsubsection{Optional MLP-weight Fusion}
\label{sec:vit_mlp_fusion}

For transformer models where the replacement layer also uses MLP
information, RepL synthesizes the two MLP linear layers from adjacent
blocks. Let $W_{\ell,1}^{\mathrm{mlp}}$ and
$W_{\ell,2}^{\mathrm{mlp}}$ denote the two MLP linear weights in block
$\ell$. The synthesized weights are
\begin{equation}
\begin{aligned}
  \widehat{W}_{r,1}^{\mathrm{mlp}}
  &=
  \frac{1}{2}
  \left(
    \overline{W}_{r-1,1}^{\mathrm{mlp}}
    +
    \overline{W}_{r+1,1}^{\mathrm{mlp}}
  \right),                                            \\
  \widehat{W}_{r,2}^{\mathrm{mlp}}
  &=
  \frac{1}{2}
  \left(
    \overline{W}_{r-1,2}^{\mathrm{mlp}}
    +
    \overline{W}_{r+1,2}^{\mathrm{mlp}}
  \right).
\end{aligned}
\label{eq:vit_mlp_weight_synthesis}
\end{equation}
The biases are averaged in the same way:
\begin{equation}
\begin{aligned}
  \widehat{b}_{r,1}^{\mathrm{mlp}}
  &=
  \frac{1}{2}
  \left(
    b_{r-1,1}^{\mathrm{mlp}}
    +
    b_{r+1,1}^{\mathrm{mlp}}
  \right),                                            \\
  \widehat{b}_{r,2}^{\mathrm{mlp}}
  &=
  \frac{1}{2}
  \left(
    b_{r-1,2}^{\mathrm{mlp}}
    +
    b_{r+1,2}^{\mathrm{mlp}}
  \right).
\end{aligned}
\label{eq:vit_mlp_bias_synthesis}
\end{equation}
The MLP replacement branch is
\begin{equation}
\begin{aligned}
  \mathbf{H}_r
  &=
  \operatorname{GELU}
  \!\left(
    \operatorname{LN}_{r-1,2}(\mathbf{X})
    \left(
      \widehat{W}_{r,1}^{\mathrm{mlp}}
    \right)^{\top}
    +
    \widehat{b}_{r,1}^{\mathrm{mlp}}
  \right),                                            \\
  \Delta_r^{\mathrm{mlp}}
  &=
  \mathbf{H}_r
  \left(
    \widehat{W}_{r,2}^{\mathrm{mlp}}
  \right)^{\top}
  +
  \widehat{b}_{r,2}^{\mathrm{mlp}},                   \\
  g_r^{\mathrm{mlp}}(\mathbf{X})
  &=
  \mathbf{X}
  +
  \Delta_r^{\mathrm{mlp}} .
\end{aligned}
\label{eq:vit_mlp_forward}
\end{equation}

\subsection{Training and Gradient Flow}
\label{sec:training}

RepL is trained end-to-end with the same task objective as the original
network. For classification, we use the cross-entropy loss
\begin{equation}
\begin{aligned}
  \mathcal{L}
  =
  -
  \sum_{c=1}^{C}
  y_c
  \log p_c .
\end{aligned}
\label{eq:ce_loss}
\end{equation}
The optimized parameters include the retained backbone parameters, the
prediction head, and the lightweight parameters introduced by the
computing layers:
\begin{equation}
\begin{aligned}
  \min_{\Theta_{\mathrm{keep}},\Theta_{\mathrm{head}},\Psi}
  \;
  \mathbb{E}_{(\mathbf{x},y)}
  \left[
    \ell
    \!\left(
      F_{\mathrm{RepL}}
      \!\left(
        \mathbf{x};
        \Theta_{\mathrm{keep}},
        \Theta_{\mathrm{head}},
        \Psi
      \right),
      y
    \right)
  \right].
\end{aligned}
\label{eq:repl_objective}
\end{equation}
Here $\Psi=\{\psi_r:r\in\mathcal{R}\}$ denotes the additional
parameters in the computing layers, such as the channel-wise or
head-wise synthesis coefficients.

In the implementation, the neighboring weights used for synthesis can
be read as fixed sources for the replacement branch:
\begin{equation}
\begin{aligned}
  \widehat{\theta}_r
  =
  \mathcal{T}^{\mathrm{prev}}_r
  \!\left(
    \operatorname{sg}(\theta_{r-1})
  \right)
  +
  \mathcal{T}^{\mathrm{next}}_r
  \!\left(
    \operatorname{sg}(\theta_{r+1})
  \right),
\end{aligned}
\label{eq:stop_gradient}
\end{equation}
where $\operatorname{sg}(\cdot)$ denotes stop-gradient. This means that the computing layer uses adjacent weights as synthesis anchors, while the retained backbone blocks are updated through their own ordinary
forward paths.

The gradients of the computing-layer parameters are obtained by the
standard chain rule:
\begin{equation}
\begin{aligned}
  \nabla_{\psi_r}\mathcal{L}
  =
  \frac{\partial \mathcal{L}}
       {\partial g_r}
  \frac{\partial g_r}
       {\partial \widehat{\theta}_r}
  \frac{\partial \widehat{\theta}_r}
       {\partial \psi_r}.
\end{aligned}
\label{eq:chain_rule}
\end{equation}
Therefore, RepL does not require a special optimizer or an auxiliary
local loss.

\subsection{Operator Summary}
\label{sec:operator_summary}

For CNN BasicBlocks, RepL physically removes the original two
$3\times3$ convolutions and replaces them with one synthesized
$3\times3$ convolution followed by BN and residual addition. For CNN
Bottlenecks, RepL physically removes the original
$1\times1$--$3\times3$--$1\times1$ block and constructs a replacement
Bottleneck-style computation using adjacent projection weights and a
synthesized the middle spatial kernel. For ViTs, RepL physically removes the original transformer block and replaces it with a lightweight
token-wise transformation synthesized from adjacent projection weights,
optionally with head-wise synthesis and MLP-weight fusion.

Overall, RepL keeps the structural continuity of the original backbone
by using neighboring retained blocks as weight-space anchors, while
reducing the cost of training every full block in the original network.

\section{Experiments}
\subsection{Experimental setup}
We conduct classification and detection experiments using different architectures on five benchmark datasets: CIFAR-10 \cite{r1}, STL-10 \cite{r3}, SVHN \cite{r2}, ImageNet \cite{r4}, and COCO \cite{r51}. All models are trained from scratch without using any pre-trained weights. We set $k=4$ as the interval for the removed layer, and all layers compute the loss using gradient descent and update parameters via backpropagation \cite{r12}.

For the small-scale datasets (CIFAR-10, SVHN, and STL-10), we employ ViT-Tiny/8 \cite{r14}, ResNet-32, and ResNet-110 \cite{r5}, with training performed on a single Nvidia A100 GPU. For the ViT models, we use a batch size of 512 and the AdamW optimizer with a learning rate of 1e-3, training for 250 epochs. For the ResNet models, we use a batch size of 1024 and the SGD optimizer with a learning rate of 0.8, also training for 250 epochs. The data augmentation strategies are as follows: on CIFAR-10, 4-pixel reflection padding followed by random cropping back to 32$\times$32 and horizontal flipping with probability 0.5; on SVHN, random cropping to 32$\times$32 (with 2-pixel padding) without horizontal flipping; on STL-10, random cropping to 96$\times$96 (with 4-pixel padding) and horizontal flipping with probability 0.5.

For ImageNet, we conduct experiments on 4 Nvidia A100 GPUs. ViT-Tiny/16 and ViT-Small/16 are trained with a batch size of 1024 and the AdamW optimizer at a learning rate of 7.5e-4, while ResNet-34, ResNet-101, and ResNet-152 are trained with a batch size of 512 and the SGD optimizer at a learning rate of 0.2, for 90 epochs. During training, we apply a 224$\times$224 random crop with random horizontal flipping; at test time, images are resized and then center-cropped to 224$\times$224.

\subsection{Comparison with the End-to-End (E2E) results}
\subsubsection{\textbf{Results on CIFAR-10, SVHN, and STL-10}}

\begin{table*}[t]
\caption{Performance of different backbones on various datasets. RepL represents Replacement Learning. }
\label{t1}
\centering
\scalebox{1.0}{%
\setlength{\tabcolsep}{5pt}
\setlength{\tabcolsep}{5.0mm}
\begin{tabular}{cccccc}
\toprule
\textbf{Dataset} & \textbf{Backbone} & \textbf{Method} &
\textbf{Test Accuracy (\%)} & \textbf{GPU Memory (GB)} &
\textbf{Training Time (s/epoch)} \\ \midrule

\multirow{7}{*}{CIFAR-10}
& \multirow{2}{*}{ResNet-32} & E2E
& 93.17$\pm$0.14 & 3.38 & 10.44 \\
&                               & \cellcolor[HTML]{E6F0FF}RepL
& \cellcolor[HTML]{E6F0FF}93.43$\pm$0.19 ($\uparrow$0.26)
& \cellcolor[HTML]{E6F0FF}2.69 ($\downarrow$20.4\%)
& \cellcolor[HTML]{E6F0FF}8.20 ($\downarrow$21.5\%) \\ \cline{2-6}
& \multirow{2}{*}{ResNet-110} & E2E
& 93.49$\pm$0.29 & 9.31 & 26.19 \\
&                               & \cellcolor[HTML]{E6F0FF}RepL
& \cellcolor[HTML]{E6F0FF}94.01$\pm$0.17 ($\uparrow$0.52)
& \cellcolor[HTML]{E6F0FF}7.62 ($\downarrow$18.2\%)
& \cellcolor[HTML]{E6F0FF}20.93 ($\downarrow$20.1\%) \\ \cline{2-6}
& \multirow{3}{*}{ViT-Tiny/8} & E2E
& 72.77$\pm$1.31 & 2.81 & 6.81 \\
&                               & Skip-Attention
& 72.60$\pm$3.57($\downarrow$0.17) & 2.12($\downarrow$24.6\%) & 6.23($\downarrow$8.5\%) \\
&                               & \cellcolor[HTML]{E6F0FF}RepL
& \cellcolor[HTML]{E6F0FF}73.71$\pm$1.08 ($\uparrow$0.94)
& \cellcolor[HTML]{E6F0FF}2.08 ($\downarrow$26.0\%)
& \cellcolor[HTML]{E6F0FF}5.65 ($\downarrow$17.0\%) \\ \midrule

\multirow{7}{*}{SVHN}
& \multirow{2}{*}{ResNet-32} & E2E
& 96.83$\pm$0.15 & 3.38 & 13.89 \\
&                               & \cellcolor[HTML]{E6F0FF}RepL
& \cellcolor[HTML]{E6F0FF}96.97$\pm$0.12 ($\uparrow$0.14)
& \cellcolor[HTML]{E6F0FF}2.69 ($\downarrow$20.4\%)
& \cellcolor[HTML]{E6F0FF}11.94 ($\downarrow$14.0\%) \\ \cline{2-6}
& \multirow{2}{*}{ResNet-110} & E2E
& 96.93$\pm$0.24 & 9.31 & 37.38 \\
&                               & \cellcolor[HTML]{E6F0FF}RepL
& \cellcolor[HTML]{E6F0FF}97.06$\pm$0.27 ($\uparrow$0.13)
& \cellcolor[HTML]{E6F0FF}7.62 ($\downarrow$18.2\%)
& \cellcolor[HTML]{E6F0FF}30.08 ($\downarrow$19.5\%) \\ \cline{2-6}
& \multirow{3}{*}{ViT-Tiny/8} & E2E
& 85.99$\pm$0.71 & 2.81 & 10.07 \\
&                               & Skip-Attention
& 86.22$\pm$1.51($\uparrow$0.23) & 2.12($\downarrow$24.6\%) & 9.18($\downarrow$8.8\%) \\
&                               & \cellcolor[HTML]{E6F0FF}RepL
& \cellcolor[HTML]{E6F0FF}86.67$\pm$1.18 ($\uparrow$0.68)
& \cellcolor[HTML]{E6F0FF}2.08 ($\downarrow$26.0\%)
& \cellcolor[HTML]{E6F0FF}8.08 ($\downarrow$19.8\%) \\ \midrule

\multirow{7}{*}{STL-10}
& \multirow{2}{*}{ResNet-32} & E2E
& 79.81$\pm$0.51 & 3.38 & 5.11 \\
&                               & \cellcolor[HTML]{E6F0FF}RepL
& \cellcolor[HTML]{E6F0FF}80.33$\pm$0.42 ($\uparrow$0.52)
& \cellcolor[HTML]{E6F0FF}2.69 ($\downarrow$20.4\%)
& \cellcolor[HTML]{E6F0FF}4.13 ($\downarrow$19.2\%) \\ \cline{2-6}
& \multirow{2}{*}{ResNet-110} & E2E
& 79.78$\pm$0.30 & 9.31 & 6.86 \\
&                               & \cellcolor[HTML]{E6F0FF}RepL
& \cellcolor[HTML]{E6F0FF}80.45$\pm$0.51 ($\uparrow$0.67)
& \cellcolor[HTML]{E6F0FF}7.62 ($\downarrow$18.2\%)
& \cellcolor[HTML]{E6F0FF}5.23 ($\downarrow$23.8\%) \\ \cline{2-6}
& \multirow{3}{*}{ViT-Tiny/8} & E2E
& 49.08$\pm$3.39 & 2.81 & 2.93 \\
&                               & Skip-Attention
& 50.42$\pm$3.18($\uparrow$1.34) & 2.12($\downarrow$24.6\%) & 2.68($\downarrow$8.5\%) \\
&                               & \cellcolor[HTML]{E6F0FF}RepL
& \cellcolor[HTML]{E6F0FF}50.66$\pm$3.18 ($\uparrow$1.58)
& \cellcolor[HTML]{E6F0FF}2.08 ($\downarrow$26.0\%)
& \cellcolor[HTML]{E6F0FF}2.41 ($\downarrow$17.8\%) \\
\bottomrule
\end{tabular}}
\end{table*}
We evaluate our method on CIFAR-10 \cite{r1}, SVHN \cite{r2}, and STL-10 \cite{r3} and present the results in Table~\ref{t1}.
The results reveal that Replacement Learning (RepL) consistently outperforms E2E training \cite{r12} across all architectures: On CIFAR-10 \cite{r1}, ResNet-32/110 \cite{r5} test accuracy rises from 93.17 to 93.43 and 93.49 to 94.01, while ViT-Tiny/8 \cite{r14} gains 0.94; on SVHN \cite{r2}, accuracy increases by 0.13 at least across networks; on STL-10 \cite{r3}, gains range from 0.52 to 1.58, with consistent significant improvements across datasets. Table~\ref{t1} also shows RepL’s advantages on CIFAR-10 \cite{r1}: ResNet-32/110 \cite{r5}, and ViT-Tiny/8 \cite{r14} reduce GPU memory by 0.69/1.69/0.73 GB, and training time per epoch by 21.5\%, 20.1\%, 17.0\% respectively. Similar trends hold for SVHN \cite{r2} and STL-10 \cite{r3}, where RepL cuts memory and training time while maintaining or improving performance.

Furthermore, when compared to Skip-Attention \cite{r21} on ViTs \cite{r14}, our method outperforms both in terms of performance and resource efficiency, making it a more favorable choice for maintaining accuracy while reducing computational cost.
\subsubsection{\textbf{Results on ImageNet}}
\begin{table*}[t]
\caption{Results on the ImageNet validation set. RepL stands for Replacement Learning.}
\label{t2}
\centering
\scalebox{1.0}{%
\setlength{\tabcolsep}{5pt}
\setlength{\tabcolsep}{5.4mm}
\begin{tabular}{cccccc}
\toprule
\textbf{Backbone} & \textbf{Method} &
\textbf{\begin{tabular}[c]{@{}c@{}}Top-1\\Accuracy (\%)\end{tabular}} &
\textbf{\begin{tabular}[c]{@{}c@{}}Top-5\\Accuracy (\%)\end{tabular}} &
\textbf{\begin{tabular}[c]{@{}c@{}}GPU Memory\\(GB)\end{tabular}} &
\textbf{\begin{tabular}[c]{@{}c@{}}Training Time\\(s/epoch)\end{tabular}} \\ \midrule
\multirow{2}{*}{ResNet-34}
& E2E & 74.82$\pm$1.43 & 91.04$\pm$1.33 & 9.21 & 463.23 \\
& \cellcolor[HTML]{E6F0FF}RepL
  & \cellcolor[HTML]{E6F0FF}75.44$\pm$1.27 ($\uparrow$0.62)
  & \cellcolor[HTML]{E6F0FF}91.47$\pm$2.01 ($\uparrow$0.43)
  & \cellcolor[HTML]{E6F0FF}8.06 ( $\downarrow$12.5\% )
  & \cellcolor[HTML]{E6F0FF}410.53 ( $\downarrow$11.4\% ) \\ \cline{1-6}
\multirow{2}{*}{ResNet-101}
& E2E & 77.55$\pm$1.22 & 93.80$\pm$1.78 & 20.95 & 720.11 \\
& \cellcolor[HTML]{E6F0FF}RepL
  & \cellcolor[HTML]{E6F0FF}78.13$\pm$1.65 ($\uparrow$0.58)
  & \cellcolor[HTML]{E6F0FF}94.02$\pm$1.34 ($\uparrow$0.22)
  & \cellcolor[HTML]{E6F0FF}18.05 ( $\downarrow$13.8\% )
  & \cellcolor[HTML]{E6F0FF}616.23 ( $\downarrow$14.4\% ) \\ \cline{1-6}
\multirow{2}{*}{ResNet-152}
& E2E & 78.16$\pm$1.56 & 94.03$\pm$1.25 & 27.58 & 738.74 \\
& \cellcolor[HTML]{E6F0FF}RepL
  & \cellcolor[HTML]{E6F0FF}78.31$\pm$1.46 ($\uparrow$0.15)
  & \cellcolor[HTML]{E6F0FF}94.14$\pm$1.14 ($\uparrow$0.11)
  & \cellcolor[HTML]{E6F0FF}24.19 ( $\downarrow$12.3\% )
  & \cellcolor[HTML]{E6F0FF}633.89 ( $\downarrow$14.2\% ) \\ \cline{1-6}
\multirow{3}{*}{ViT-T/16}
& E2E & 60.23$\pm$1.52 & 82.38$\pm$1.32 & 12.17 & 357.66 \\
& Skip-Attn
  & 60.51$\pm$1.20($\uparrow$0.28) & 82.72$\pm$1.09($\uparrow$0.34)
  & 11.52 ( $\downarrow$5.3\% )
  & 381.44 ( $\uparrow$6.7\% ) \\
& \cellcolor[HTML]{E6F0FF}RepL
  & \cellcolor[HTML]{E6F0FF}60.93$\pm$1.19 ($\uparrow$0.70)
  & \cellcolor[HTML]{E6F0FF}82.88$\pm$1.07 ($\uparrow$0.50)
  & \cellcolor[HTML]{E6F0FF}9.59 ( $\downarrow$21.2\% )
  & \cellcolor[HTML]{E6F0FF}290.15 ( $\downarrow$18.9\% ) \\ \cline{1-6}
\multirow{3}{*}{ViT-S/16}
& E2E & 64.35$\pm$1.83 & 84.64$\pm$1.22 & 21.05 & 798.61 \\
& Skip-Attn
  & 61.65$\pm$1.25($\downarrow$2.70) & 82.70$\pm$1.16($\downarrow$1.94)
  & 20.67 ( $\downarrow$1.8\% )
  & 755.14 ( $\downarrow$5.4\% ) \\
& \cellcolor[HTML]{E6F0FF}RepL
  & \cellcolor[HTML]{E6F0FF}65.09$\pm$1.41 ($\uparrow$0.74)
  & \cellcolor[HTML]{E6F0FF}85.42$\pm$1.73 ($\uparrow$0.78)
  & \cellcolor[HTML]{E6F0FF}16.22 ( $\downarrow$22.9\% )
  & \cellcolor[HTML]{E6F0FF}617.10 ( $\downarrow$22.7\% ) \\
\midrule
\multirow{3}{*}{ViT-B/16}
& E2E & 59.46$\pm$1.72 & 80.35$\pm$1.12 & 41.97 & 2566.70 \\
& Skip-Attn
  & 58.94$\pm$1.25($\downarrow$0.52) & 79.70$\pm$0.94($\downarrow$0.65)
  & 38.49 ( $\downarrow$8.3\% )
  & 2393.81 ( $\downarrow$6.7\% ) \\
& \cellcolor[HTML]{E6F0FF}RepL
  & \cellcolor[HTML]{E6F0FF}60.18$\pm$1.27 ($\uparrow$0.72)
  & \cellcolor[HTML]{E6F0FF}81.97$\pm$1.15 ($\uparrow$1.62)
  & \cellcolor[HTML]{E6F0FF}29.94 ( $\downarrow$28.7\% )
  & \cellcolor[HTML]{E6F0FF}1924.35 ( $\downarrow$25.1\% ) \\
\bottomrule
\end{tabular}}
\end{table*}
%
We further evaluate RepL on ImageNet \cite{r4}, a substantially larger and more diverse benchmark than CIFAR-10, SVHN, and STL-10, using ResNet-34/101/152 \cite{r5}, and ViT-Tiny/16, ViT-Small/16, and ViT-Base/16 \cite{r14}; the results are summarized in Table~\ref{t2}. Across all six backbones, RepL consistently improves both Top-1 and Top-5 accuracy over E2E training, with Top-1 gains of 0.62, 0.58, 0.15, 0.70, 0.74, and 0.72 points, respectively. This consistency is important because ImageNet requires stronger semantic discrimination and richer feature hierarchies than the small-scale datasets. The gains, therefore, indicate that replacing selected blocks with synthesized operators does not merely work as a regularizer on simple datasets, but can preserve and even improve representation learning under a large-scale visual recognition setting.

The magnitude of the improvement also varies with model family. For ResNets, the gain is more visible on ResNet-34 and ResNet-101 than on ResNet-152, suggesting that deeper CNNs already contain more redundant capacity and leave less room for accuracy improvement, while still benefiting from reduced computation. For ViTs, RepL yields larger Top-1 gains and a particularly strong Top-5 gain on ViT-Base/16. This pattern is aligned with the motivation of RepL: transformer blocks contain highly correlated attention and MLP transformations across neighboring layers, and synthesizing a replacement operator from adjacent blocks can remove redundant computation while maintaining contextual continuity.

In terms of efficiency, RepL reduces GPU memory by 12.3\%-28.7\% and training time by 11.4\%-25.1\% across the ImageNet experiments. These savings directly reflect the design of RepL: the removed full block no longer needs to store all intermediate activations for backpropagation, while the lightweight replacement block only introduces a small parameter-fusion computation. The ViT results further clarify the difference from Skip-Attention \cite{r21}. Skip-Attention occasionally improves accuracy on ViT-Tiny/16, but it increases training time and becomes unstable on larger ViTs, dropping 2.70 Top-1 points on ViT-Small/16 and 0.52 points on ViT-Base/16. In contrast, RepL improves accuracy and reduces both memory and time on all ViT backbones, showing that learning a replacement operator from neighboring parameters is more robust than directly reusing attention computations.

\subsection{Ablation study}
\subsubsection{\textbf{Performance analysis of computing layer usage}}
\begin{table}[t]
\renewcommand{\arraystretch}{0.8}
\caption{Performance comparison on CIFAR-10.}
\label{t3}
\setlength{\tabcolsep}{5pt}
\centering
\scalebox{0.9}{
\setlength{\tabcolsep}{2.0mm}
\begin{tabular}{ccccc}
\toprule
\textbf{Backbone}
& \textbf{Method}
& \textbf{\makecell{Test\\Accuracy\\(\%)}}
& \textbf{\makecell{GPU\\Memory\\(GB)}}
& \textbf{\makecell{Training Time\\(s/epoch)}} \\
\midrule
                             & E2E                        & 93.49±0.29          & 9.31               & 26.19            \\
                             & - 25\% layers              & 92.62±2.01          & 7.07               & 19.54            \\
\multirow{-3}{*}{ResNet-110}  & \cellcolor[HTML]{E6F0FF}+ computing layers & \cellcolor[HTML]{E6F0FF}94.01±0.17 & \cellcolor[HTML]{E6F0FF}7.62 & \cellcolor[HTML]{E6F0FF}20.93 \\
\midrule
                             & E2E                        & 72.77±1.31          & 2.81               & 6.81             \\
                             & - 25\% layers              & 71.13±1.24          & 2.04               & 5.44             \\
\multirow{-3}{*}{ViT-Tiny/8} & \cellcolor[HTML]{E6F0FF}+ computing layers & \cellcolor[HTML]{E6F0FF}73.71±1.08 & \cellcolor[HTML]{E6F0FF}2.08 & \cellcolor[HTML]{E6F0FF}5.65  \\
\bottomrule
\end{tabular}}
\end{table}

\begin{table}[t]
\renewcommand{\arraystretch}{0.8}
\caption{Performance comparison on ImageNet.}
\label{t4}
\centering
\scalebox{0.90}{
\setlength{\tabcolsep}{1.1mm}
\begin{tabular}{cccccc}
\toprule
\textbf{Backbone}
& \textbf{Method}
& \textbf{\makecell{Top-1\\Accuracy\\(\%)}}
& \textbf{\makecell{Top-5\\Accuracy\\(\%)}}
& \textbf{\makecell{GPU\\Memory\\(GB)}}
& \textbf{\makecell{Training\\Time\\(s/epoch)}} \\
\midrule
& E2E & 74.82±1.43 & 91.04±1.33 & 9.21 & 463.23 \\
& - 25\% layers & 72.99±1.82 & 90.12±1.31 & 7.75 & 392.21 \\
\multirow{-3}{*}{ResNet-34} & \cellcolor[HTML]{E6F0FF}+ computing layers & \cellcolor[HTML]{E6F0FF}75.44±1.27 & \cellcolor[HTML]{E6F0FF}91.47±2.01 & \cellcolor[HTML]{E6F0FF}8.06 & \cellcolor[HTML]{E6F0FF}410.53 \\
\midrule
& E2E & 60.23±1.52 & 82.38±1.33 & 12.17 & 357.66 \\
& - 25\% layers & 58.22±0.91 & 81.51±1.22 & 9.49 & 287.55 \\
\multirow{-3}{*}{ViT-Tiny/16} & \cellcolor[HTML]{E6F0FF}+ computing layers & \cellcolor[HTML]{E6F0FF}60.93±1.19 & \cellcolor[HTML]{E6F0FF}82.88±1.07 & \cellcolor[HTML]{E6F0FF}9.59 & \cellcolor[HTML]{E6F0FF}290.15 \\
\bottomrule
\end{tabular}}
\end{table}

\begin{table}[!t]
\renewcommand{\arraystretch}{0.9} 
\caption{Performance comparison on CIFAR-10 with different $k$ setting.}
\label{t5}
\centering
\scalebox{0.9}{
\setlength{\tabcolsep}{2.3mm}
\begin{tabular}{ccccc}
\toprule
\textbf{Backbone}
& \textbf{$k$ value setting}
& \textbf{\makecell{Test\\Accuracy\\(\%)}}
& \textbf{\makecell{GPU\\Memory\\(GB)}}
& \textbf{\makecell{Training Time\\(s/epoch)}} \\
\midrule
& $k=2$ & 92.67±1.89 & 6.25 & 18.05 \\
& \cellcolor[HTML]{E6F0FF}$k=4$ & \cellcolor[HTML]{E6F0FF}94.01±0.17 & \cellcolor[HTML]{E6F0FF}7.62 & \cellcolor[HTML]{E6F0FF}20.93 \\
\multirow{-3}{*}{ResNet-110} & $k=6$ & 94.11±0.34 & 8.63 & 23.94 \\
\midrule
& $k=2$ & 71.48±2.39 & 1.70 & 5.19 \\
& \cellcolor[HTML]{E6F0FF}$k=4$ & \cellcolor[HTML]{E6F0FF}73.71±1.08 & \cellcolor[HTML]{E6F0FF}2.08 & \cellcolor[HTML]{E6F0FF}5.65 \\
\multirow{-3}{*}{ViT-Tiny/8} & $k=6$ & 73.94±1.17 & 2.39 & 6.39 \\
\bottomrule
\end{tabular}}
\end{table}

\begin{table}[!t]
\renewcommand{\arraystretch}{0.9} 
\caption{Ablation of Parameters in Computing Layers.}
\label{tab:repl_ablation}
\centering
\scalebox{0.9}{
\setlength{\tabcolsep}{2.5mm}
\begin{tabular}{lccc}
\toprule
\textbf{Method} 
& \textbf{\makecell{Accuracy\\(\%)}} 
& \textbf{\makecell{GPU\\Memory\\(GB)}} 
& \textbf{\makecell{Training Time\\(sec / epoch)}} \\
\midrule
\cellcolor[HTML]{E6F0FF}RepL & \cellcolor[HTML]{E6F0FF}73.71±1.08 & \cellcolor[HTML]{E6F0FF}2.08 & \cellcolor[HTML]{E6F0FF}5.65 \\
RepL (only Attention weights) & 72.39±0.97 & 2.05 & 5.59 \\
RepL (only MLP weights) & 72.14±1.34 & 2.07 & 5.53 \\
RepL (no weights) & 69.30±2.11 & 2.02 & 5.20 \\
\bottomrule
\end{tabular}}
\end{table}

%
To examine whether the gain comes from simple layer removal or from the replacement design, we compare E2E training \cite{r12}, direct removal of 25\% layers with \(k=4\), and RepL with computing layers. The evaluation is conducted on CIFAR-10 \cite{r1} with ViT-Tiny/8 \cite{r14} and ResNet-110 \cite{r5}, and on ImageNet \cite{r4} with ViT-Tiny/16 \cite{r14} and ResNet-34 \cite{r5}. This comparison separates the effect of reducing executed depth from the effect of reconstructing the removed transformation with neighboring parameters.

As shown in Table~\ref{t3} and \ref{t4}, direct layer removal reduces memory and training time but lowers accuracy on both CNN and ViT backbones, which suggests that the removed blocks still provide useful intermediate transformations. After adding computing layers, RepL recovers the lost accuracy and further improves over E2E training with only a small extra cost over direct removal. This result shows that the computing layer is the key component that turns layer removal into an effective replacement strategy.

\subsubsection{\textbf{Analysis of interval setting for removed layers}}
%
We study the interval \(k\) because it controls how frequently blocks are replaced and therefore determines the balance between representation preservation and resource saving. We compare \(k=2\), \(k=4\), and \(k=6\) on CIFAR-10 \cite{r1} using ViT-Tiny/8 \cite{r14} and ResNet-110 \cite{r5}. A smaller \(k\) replaces blocks more aggressively, while a larger \(k\) keeps more of the original network.

As shown in Table~\ref{t5}, \(k=2\) gives stronger memory and time reduction but causes clear accuracy degradation, indicating that overly frequent replacement weakens the continuity of feature transformation. In contrast, \(k=6\) preserves more original blocks and obtains slightly better accuracy, but the efficiency gain becomes weaker. Therefore, \(k=4\) offers a practical balance by retaining enough neighboring context for reliable replacement while still removing enough full blocks to reduce cost.

\subsubsection{\textbf{Comparison of features in different methods}}
\begin{figure*}[!htbp]
\begin{minipage}[b]{0.245\linewidth}
  \centering
  \includegraphics[width=2.4cm]{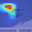}
  \centerline{(a)}\medskip
\end{minipage}
\begin{minipage}[b]{0.245\linewidth}
  \centering
  \includegraphics[width=2.4cm]{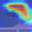}
  \centerline{(b)}\medskip
\end{minipage}
\begin{minipage}[b]{0.245\linewidth}
  \centering
  \includegraphics[width=2.4cm]{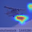}
  \centerline{(c)}\medskip
\end{minipage}
\begin{minipage}[b]{0.245\linewidth}
  \centering
  \includegraphics[width=2.4cm]{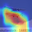}
  \centerline{(d)}\medskip
\end{minipage}
\caption{Visualization of feature maps. (a) Feature map of ResNet-32 with End-to-End training. (b) Feature map of ResNet-32 with Replacement Learning. (c) Feature map of ViT-Tiny/8 with End-to-End training. (d) Feature map of ViT-Tiny/8 with Replacement Learning.} 
\label{f3} 
\end{figure*}

\begin{table*}[t]
\renewcommand{\arraystretch}{0.8} 
\caption{Performance comparison on CIFAR-10 using different layers.}
\label{t6}
\centering
\scalebox{1.0}{
\setlength{\tabcolsep}{7.2mm}
\begin{tabular}{ccc|ccc}
\toprule
\multicolumn{3}{c|}{ResNet-110}                                                                                                               & \multicolumn{3}{c}{ViT-Tiny/8}                                                                                                                \\ \midrule
\begin{tabular}[c]{@{}c@{}}Preceding\\ Layer\end{tabular} & \begin{tabular}[c]{@{}c@{}}Succeeding\\ Layer\end{tabular} & Test Accuracy (\%)  & \begin{tabular}[c]{@{}c@{}}Preceding\\ Layer\end{tabular} & \begin{tabular}[c]{@{}c@{}}Succeeding\\ Layer\end{tabular} & Test Accuracy (\%)  \\ \midrule
\ding{55}                                                          & \ding{55}                                                          & 93.49±0.29          & \ding{55}                                                           & \ding{55}                                                           & 72.77±1.31          \\
\ding{55}                                                           & \ding{51}                                                           & 93.19±1.38          & \ding{55}                                                           & \ding{51}                                                          & 72.18±1.93          \\
\ding{51}                                                          & \ding{55}                                                           & 91.75±2.31          & \ding{51}                                                          & \ding{55}                                                           & 69.37±4.85          \\ \rowcolor[HTML]{E6F0FF}
\textbf{\ding{51}}                                                 & \textbf{\ding{51}}                                                 & \textbf{94.01±0.17} & \textbf{\ding{51}}                                                 & \textbf{\ding{51}}                                                 & \textbf{73.71±1.08} \\ \bottomrule
\end{tabular}}
\end{table*}

%
We further visualize feature maps on CIFAR-10 \cite{r1} with ResNet-32 \cite{r5} to examine how RepL affects the learned representations. As shown in Figure~\ref{f3}, the E2E feature maps in (a) and (c) are concentrated in a limited set of regions, while the RepL feature maps in (b) and (d) activate broader spatial areas and preserve clearer local structures.

This qualitative difference suggests that the replacement blocks do not simply remove information from the network. Instead, by combining cues from neighboring layers, RepL encourages the model to retain both local details and wider contextual responses. This observation is consistent with the quantitative improvements in the previous tables, where RepL improves accuracy while reducing the cost of full block execution.

\subsubsection{\textbf{Comparison of using different parts of parameters}}
To validate the importance of using neighboring parameters in the computing layer, we conduct an ablation study on CIFAR-10 with ViT-Tiny/8. We compare four configurations, including the full RepL design that uses both attention and MLP parameters, a variant using only attention parameters, a variant using only MLP parameters, and a variant that does not use adjacent parameters.

As summarized in Table~\ref{tab:repl_ablation}, the full design achieves the best accuracy, while removing either attention or MLP information leads to lower performance. The drop is largest when no adjacent parameters are used, which shows that the computing layer needs structural information from neighboring blocks rather than only an extra learnable mapping. The stronger result from attention parameters also indicates that attention carries more transferable contextual information for replacing a ViT block.

\subsubsection{\textbf{Comparison of using different layers}}
%
We also examine whether the computing layer should use information from one neighbor or from both neighbors. Using ResNet-110 \cite{r5} and ViT-Tiny/8 \cite{r14} on CIFAR-10 \cite{r1}, we compare variants that use only the preceding layer, only the succeeding layer, and both layers.

As shown in Table~\ref{t6}, using only one side leads to a clear accuracy drop, while using both preceding and succeeding layers achieves the best performance and surpasses E2E training \cite{r12}. This result supports the core design of RepL. The preceding layer provides historical feature context, while the succeeding layer provides information about the target transformation, and their combination gives the computing layer a more reliable basis for replacing the removed block.

\subsection{Analytical experiments}
\label{sec:analytical_experiments}

\subsubsection{\textbf{Empirical analysis}}

\paragraph{Detection experiments and analysis}
To evaluate whether RepL generalizes beyond image classification, we
conduct object detection experiments on COCO~\cite{r51} using
RetinaNet-R50 and RetinaNet-R101~\cite{r52} as backbones. We use
4 NVIDIA A100 GPUs, a batch size of 8, the Adam optimizer, a learning
rate of $4\times 10^{-5}$, and train all models for 100 epochs. The
results are reported in Table~\ref{tab:coco_detection}.

\begin{table*}[!t]
\renewcommand{\arraystretch}{0.8}
\caption{Performance comparison on COCO using different backbones. * means the addition of Replacement Learning.}
\label{tab:coco_detection}
\centering
\scalebox{1.0}{
\setlength{\tabcolsep}{4.9mm}
\begin{tabular}{cccccc}
\toprule
\textbf{Backbone}
& \textbf{mAP}
& \textbf{AP@50}
& \textbf{AP@75}
& \textbf{GPU Memory (GB)}
& \textbf{Training Time (s/epoch)} \\
\midrule
RetinaNet-R50
& 30.42
& 51.72
& 30.80
& 6.85
& 3859.11 \\
\rowcolor[HTML]{E6F0FF}
RetinaNet-R50*
& 30.64($\uparrow$0.22)
& 52.44($\uparrow$0.72)
& 31.15($\uparrow$0.35)
& 5.82($\downarrow$15.04\%)
& 3245.23($\downarrow$15.91\%) \\
RetinaNet-R101
& 32.36
& 54.21
& 32.91
& 8.19
& 5548.09 \\
\rowcolor[HTML]{E6F0FF}
RetinaNet-R101*
& 32.76($\uparrow$0.40)
& 54.80($\uparrow$0.59)
& 32.98($\uparrow$0.07)
& 6.65($\downarrow$18.80\%)
& 4671.33($\downarrow$15.80\%) \\
\bottomrule
\end{tabular}}
\end{table*}

Table~\ref{tab:coco_detection} shows that RepL improves detection
accuracy while reducing training cost. On RetinaNet-R50, RepL improves
mAP from 30.42 to 30.64 and AP@50 from 51.72 to 52.44, while reducing
GPU memory by 15.04\% and training time by 15.91\%. On the deeper
RetinaNet-R101, RepL further improves mAP by 0.40 and reduces memory by
18.80\%. The gain at AP@75 is smaller for R101, suggesting that the main improvement is not merely high-IoU localization, but more efficient
feature learning and stronger general object responses. These results
indicate that RepL can be inserted into detection backbones without
disrupting dense prediction pipelines.

\begin{figure}[h]
\begin{center}
\begin{minipage}[b]{0.49\linewidth}
  \centering
  \includegraphics[width=4.5cm]{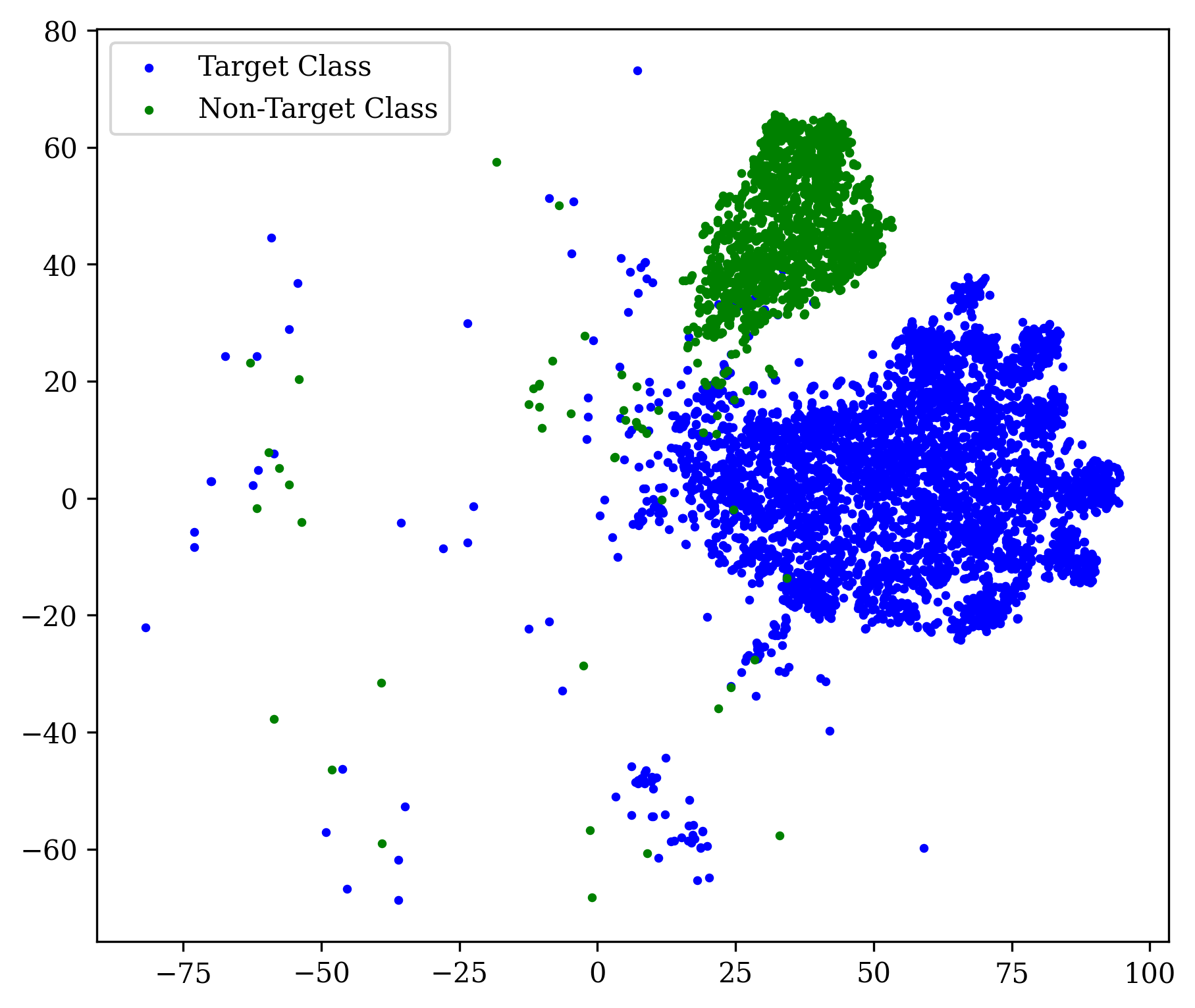}
  \centerline{(a)}\medskip
\end{minipage}
\begin{minipage}[b]{0.49\linewidth}
  \centering
  \includegraphics[width=4.5cm]{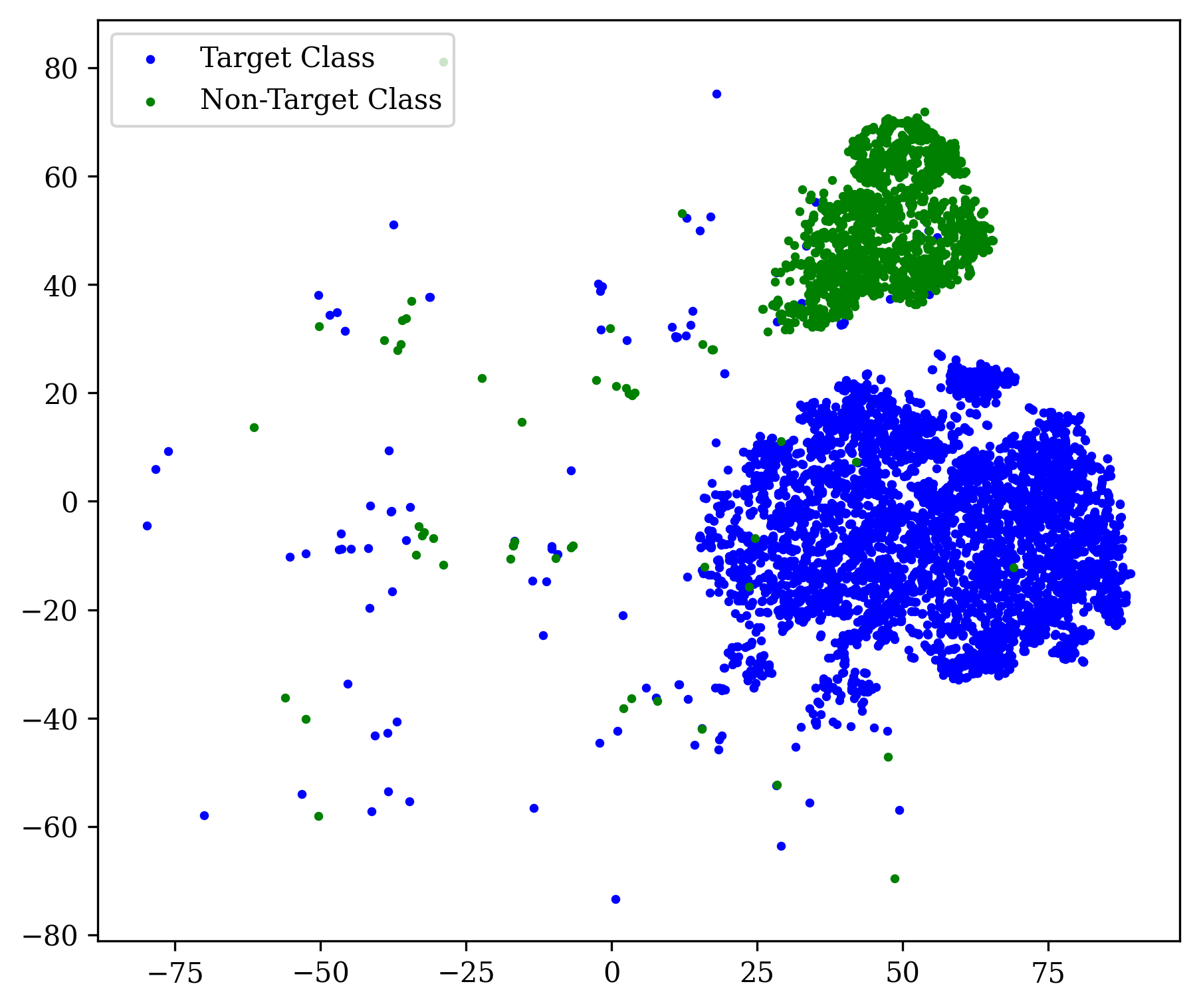}
  \centerline{(b)}\medskip
\end{minipage}
\end{center}
\caption{T-SNE visualization. (a) is t-SNE of E2E training, and (b) is t-SNE of Replacement Learning.}
\label{fig:tsne}
\end{figure}

\paragraph{Comparison of the distribution of classified data points}
To examine the learned feature geometry, we visualize the feature
distribution of ResNet-110 on SVHN~\cite{r2} using t-SNE~\cite{r50}.
The visualization is shown in Figure~\ref{fig:tsne}.

In Figure~\ref{fig:tsne} (a), the E2E features show visible overlap
between several class clusters, indicating that the final representation
still contains inter-class confusion. In contrast, Figure~\ref{fig:tsne} (b)
shows that RepL produces more compact clusters with clearer boundaries.
This qualitative result is consistent with the accuracy improvements in the classification experiments: the replacement layer does not simply
reduce computation, but also encourages more discriminative
representations by combining adjacent-layer information. We note that
t-SNE is a qualitative diagnostic rather than a standalone proof; nevertheless, it provides useful evidence that RepL improves feature
organization in the embedding space.

\paragraph{Comparative experiments with related methods}
To verify whether RepL is compatible with the existing efficiency
techniques, we compare it with Stochastic Depth~\cite{sd} and
Checkpointing~\cite{checkpointing}, and further combine RepL with each
of them. The results are reported in Table~\ref{tab:repl_sd_ckpt}. Since this table is based on a single run, we focus on the overall trend
rather than small numerical fluctuations.

\begin{table}[ht]
\caption{Comparative experiments with Stochastic Depth and Checkpointing. The results are based on a single run.}
\centering
\setlength{\tabcolsep}{3pt}
\scalebox{0.9}{
\setlength{\tabcolsep}{1.0mm}
\begin{tabular}{@{}lllccc@{}}
\toprule
\textbf{Dataset}
& \textbf{Backbone}
& \textbf{Method}
& \textbf{Acc@1}
& \textbf{\makecell{GPU\\Memory\\(GB)}}
& \textbf{\makecell{Time\\(s/epoch)}} \\
\midrule
\multirow{6}{*}{CIFAR-10}
& \multirow{6}{*}{ResNet-32}
& E2E
& 93.17 & 3.38 & 10.44 \\
& & \cellcolor[HTML]{E6F0FF}RepL
& \cellcolor[HTML]{E6F0FF}93.43
& \cellcolor[HTML]{E6F0FF}2.69
& \cellcolor[HTML]{E6F0FF}8.20 \\
& & Stochastic Depth
& 93.04 & 3.31 & 10.05 \\
& & \cellcolor[HTML]{E6F0FF}RepL+Stochastic Depth
& \cellcolor[HTML]{E6F0FF}93.17
& \cellcolor[HTML]{E6F0FF}2.67
& \cellcolor[HTML]{E6F0FF}9.36 \\
& & Checkpointing
& 93.13 & 1.77 & 16.44 \\
& & \cellcolor[HTML]{E6F0FF}RepL+Checkpointing
& \cellcolor[HTML]{E6F0FF}93.24
& \cellcolor[HTML]{E6F0FF}1.64
& \cellcolor[HTML]{E6F0FF}14.54 \\
\midrule
\multirow{6}{*}{ImageNet}
& \multirow{6}{*}{ResNet-101}
& E2E
& 77.55 & 20.95 & 720 \\
& & \cellcolor[HTML]{E6F0FF}RepL
& \cellcolor[HTML]{E6F0FF}78.13
& \cellcolor[HTML]{E6F0FF}18.01
& \cellcolor[HTML]{E6F0FF}616 \\
& & Stochastic Depth
& 77.63 & 19.39 & 652 \\
& & \cellcolor[HTML]{E6F0FF}RepL+Stochastic Depth
& \cellcolor[HTML]{E6F0FF}78.11
& \cellcolor[HTML]{E6F0FF}17.12
& \cellcolor[HTML]{E6F0FF}551 \\
& & Checkpointing
& 78.25 & 14.47 & 1012 \\
& & \cellcolor[HTML]{E6F0FF}RepL+Checkpointing
& \cellcolor[HTML]{E6F0FF}78.29
& \cellcolor[HTML]{E6F0FF}12.93
& \cellcolor[HTML]{E6F0FF}819 \\
\bottomrule
\end{tabular}}
\label{tab:repl_sd_ckpt}
\end{table}

Table~\ref{tab:repl_sd_ckpt} indicates that RepL and the compared
methods reduce cost through different mechanisms. Stochastic Depth
reduces computation through random path dropping, while Checkpointing
reduces activation memory through recomputation. RepL instead
physically removes selected blocks and replaces them with lightweight
computing layers. On CIFAR-10, RepL improves accuracy over E2E while
also reducing memory and time. When combined with Checkpointing, RepL
further reduces memory from 1.77 GB to 1.64 GB and shortens the time from
16.44 to 14.54 s/epoch. On ImageNet, RepL+Stochastic Depth reaches
17.12 GB memory and 551 s/epoch, improving over Stochastic Depth alone.
These results suggest that RepL is an orthogonal component that can be
combined with other training efficiency techniques.

\paragraph{Experiments on the NLP task}
We further test RepL on language modeling using WikiText-2 with a
Transformer-LM-12L-512d-8H-2048ff model. We use basic English
tokenization, retain words with frequency at least 2, append an
\texttt{<eos>} token to each sentence, and use a BPTT length of 128.
The model is trained for 20 epochs, and the reported variance is
computed over 5 runs with different random seeds. Results are shown in
Table~\ref{tab:wikitext2}.

\begin{table*}[htbp]
\renewcommand{\arraystretch}{0.8}
\caption{Performance on WikiText-2 using Transformer-LM-12L-512d-8H-2048ff.}
\label{tab:wikitext2}
\centering
\scalebox{1.0}{
\setlength{\tabcolsep}{5.7mm}
\begin{tabular}{cclclc}
\toprule
\textbf{Dataset}
& \textbf{Model}
& \multicolumn{1}{c}{\textbf{Method}}
& \textbf{Test PPL ($\downarrow$)}
& \multicolumn{1}{c}{\textbf{\begin{tabular}[c]{@{}c@{}}GPU Memory\\ (GB)\end{tabular}}}
& \textbf{\begin{tabular}[c]{@{}c@{}}Time\\ (s/epoch)\end{tabular}} \\
\midrule
& & \multicolumn{1}{c}{E2E}
& \cellcolor[HTML]{FFFDFA}{\color[HTML]{333333} $195.42\pm1.84$}
& \multicolumn{1}{c}{\cellcolor[HTML]{FFFDFA}{\color[HTML]{333333} 10.92}}
& \cellcolor[HTML]{FFFDFA}{\color[HTML]{333333} 20.8} \\
\cline{3-6}
\multirow{-2}{*}{WikiText-2}
& \multirow{-2}{*}{\begin{tabular}[c]{@{}c@{}}Transformer-LM-12L-512d-8H-2048ff\end{tabular}}
& \multicolumn{1}{c}{\cellcolor[HTML]{E6F0FF}RepL}
& \cellcolor[HTML]{E6F0FF}{\color[HTML]{333333} $193.31\pm3.39$}
& \multicolumn{1}{c}{\cellcolor[HTML]{E6F0FF}{\color[HTML]{333333} 9.61}}
& \cellcolor[HTML]{E6F0FF}{\color[HTML]{333333} 17.7} \\
\midrule
\multicolumn{2}{c}{}
& \multicolumn{2}{l}{Hardware: Single A100}
& \multicolumn{2}{l}{Grad\_clip: 1.0} \\
\multicolumn{2}{c}{}
& \multicolumn{2}{l}{Batch size: 64}
& \multicolumn{2}{l}{Weight decay: 0.01} \\
\multicolumn{2}{c}{}
& \multicolumn{2}{l}{Optimizer: AdamW}
& \multicolumn{2}{l}{fp: 16} \\
\multicolumn{2}{c}{\multirow{-4}{*}{Configuration}}
& \multicolumn{2}{l}{Learning rate: $3\times 10^{-4}$}
& & \multicolumn{1}{l}{} \\
\bottomrule
\end{tabular}}
\end{table*}

Table~\ref{tab:wikitext2} shows that RepL reduces test perplexity from
$195.42\pm1.84$ to $193.31\pm3.39$, while reducing GPU memory from
10.92 GB to 9.61 GB and training time from 20.8 to 17.7 s/epoch. This
is important because language modeling stresses a different computation
pattern from CNNs: the dominant cost comes from token-wise projections
and sequence modeling operations rather than spatial convolutions. The
result suggests that the adjacent-layer replacement principle is not
limited to vision backbones. RepL preserves the language modeling
objective while reducing the cost of full-depth Transformer training.

\paragraph{Inference on ImageNet with deploy re-parameterization}
During training, RepL computing layers synthesize replacement operators
from adjacent retained blocks on the fly. This dynamic form is useful
for optimization, but it is not the most efficient form for inference,
because each forward pass still involves neighboring-weight access,
weight normalization, weight composition, and Python-side module lookup.
Therefore, after training, we convert each dynamic RepL computing layer
into a static deploy operator through re-parameterization.

For CNNs, the synthesized convolution kernel is precomputed once, and the BatchNorm statistics are folded into the convolution weights:
\begin{equation}
\begin{aligned}
  W_{\mathrm{deploy}}
  &=
  \frac{
    \gamma_{\mathrm{BN}}
  }{
    \sqrt{
      \sigma_{\mathrm{BN}}^2+\epsilon
    }
  }
  \widehat{W},                                               \\
  b_{\mathrm{deploy}}
  &=
  \frac{
    \gamma_{\mathrm{BN}}
  }{
    \sqrt{
      \sigma_{\mathrm{BN}}^2+\epsilon
    }
  }
  \left(
    \widehat{b}
    -
    \mu_{\mathrm{BN}}
  \right)
  +
  \beta_{\mathrm{BN}} .
\end{aligned}
\label{eq:deploy_bn_folding}
\end{equation}
After this conversion, the CNN replacement branch is executed as an
ordinary convolution followed by the original activation function, and the synthesized weights are no longer recomputed during inference.

For ViTs, the synthesized projection matrix is folded into a fixed
linear layer:
\begin{equation}
\begin{aligned}
  W_{\mathrm{deploy}}
  &=
  c_{\mathrm{scale}}
  \widehat{W},                                            \\
  b_{\mathrm{deploy}}
  &=
  c_{\mathrm{scale}}
  \widehat{b},
\end{aligned}
\label{eq:deploy_linear_folding}
\end{equation}
where $c_{\mathrm{scale}}$ denotes the fixed inference-time scaling
factor used by the replacement projection, e.g., the dimension-dependent
normalization factor. Thus, the ViT computing layer is executed as an
ordinary residual linear projection at inference time. We further build
a static deploy-only execution graph, so that the forward pass no
longer performs dictionary lookup or conditional checks for computing
layers.

\noindent\textbf{Inference modes.}
In addition to the default FP32 execution, we evaluate two standard
inference backend optimizations for ResNet34:
\begin{itemize}
  \item \textbf{Channels-last.} Both the model and input tensors are
  converted to the channels-last memory format. This changes the tensor
  memory layout used by convolution kernels, but does not change the
  model architecture, parameters, or numerical computation. It is mainly
  useful for CNNs because modern GPU convolution kernels can be faster
  under this layout.
  \item \textbf{Channels-last + compile.} On top of channels-last, we
  apply graph compilation after switching the model to deploy mode. The
  compiled graph reduces Python dispatch overhead and allows the backend
  to optimize the static inference graph. This is especially suitable
  for RepL Deploy because the dynamic computing layers have already been
  re-parameterized into ordinary static operators.
\end{itemize}
For fairness, each backend setting is applied to both the BP baseline
and the RepL Deploy model. Therefore, the reported speedups measure the
benefit of the RepL reduced architecture under the same inference
backend, rather than the benefit of an optimization applied only to RepL.

We first isolate the effect of re-parameterization itself by comparing
the dynamic RepL graph with its exact deploy version. The benchmark uses
synthetic ImageNet tensors on a single NVIDIA A100-SXM4-80GB GPU. CUDA
synchronization is used during timing, and the reported value is the median
latency. The results are shown in Table~\ref{tab:dynamic_vs_deploy}.

\begin{table}[t]
\centering
\caption{Dynamic RepL vs. exact deploy re-parameterization. The exact deploy model precomputes the synthesized operators and uses a static execution graph.}
\label{tab:dynamic_vs_deploy}
\renewcommand{\arraystretch}{0.9}
\setlength{\tabcolsep}{2.7pt}
\begin{tabular}{lcccc}
\toprule
\textbf{Model / Setting}
& \textbf{Batch}
& \textbf{\makecell{Dynamic RepL\\Latency}}
& \textbf{\makecell{Exact Deploy\\Latency}}
& \textbf{Speedup} \\
\midrule
ResNet34 RepL
& 64
& 7.861 ms/batch
& 7.858 ms/batch
& 1.000$\times$ \\
ViT-S RepL
& 64
& 37.628 ms/batch
& 37.404 ms/batch
& 1.006$\times$ \\
ViT-S RepL
& 1
& 4.381 ms/img
& 3.847 ms/img
& 1.139$\times$ \\
ViT-S RepL + AMP
& 64
& 12.103 ms/batch
& 11.860 ms/batch
& 1.020$\times$ \\
ResNet101 RepL
& 32
& 15.062 ms/batch
& 14.868 ms/batch
& 1.013$\times$ \\
ResNet101 RepL
& 1
& 9.237 ms/img
& 8.368 ms/img
& 1.104$\times$ \\
\bottomrule
\end{tabular}
\end{table}

Table~\ref{tab:dynamic_vs_deploy} shows that the exact deploy mainly removes
runtime composition overhead. For large-batch throughput, the gain is
small because most latency still comes from the retained backbone
blocks and the static replacement Conv/Linear operators. However, for
small-batch inference, where Python overhead and dynamic graph construction are more visible, deploying re-parameterization gives clearer
speedups, e.g., 1.139$\times$ on ViT-S with batch size 1 and
1.104$\times$ on ResNet101 with batch size 1. The conversion is
numerically equivalent up to floating-point precision: the CNN deploy
path matches the dynamic path exactly, while the ViT FP32 difference is
at the $10^{-6}$ level.

We then compare the full BP baseline with RepL after deploy
re-parameterization. This comparison reflects the practical inference
benefit of RepL, because the deployed model executes the physically
reduced architecture with static replacement operators. Results are
reported in Table~\ref{tab:bp_vs_repl_deploy}.

\begin{table*}[t]
\centering
\caption{Inference comparison between the full BP baseline and RepL deploy models on ImageNet synthetic inputs. Each row applies the same inference backend to both BP and RepL Deploy.}
\label{tab:bp_vs_repl_deploy}
\renewcommand{\arraystretch}{1.05}
\setlength{\tabcolsep}{5.5pt}
\begin{tabular}{lccc}
\toprule
\textbf{Model / Inference Backend}
& \textbf{\makecell{BP Baseline\\Latency}}
& \textbf{\makecell{RepL Deploy\\Latency}}
& \textbf{Speedup} \\
\midrule
ResNet34 FP32
& 9.931 ms/batch
& 7.858 ms/batch
& 1.264$\times$ \\
ResNet34 channels-last
& 9.401 ms/batch
& 7.385 ms/batch
& 1.273$\times$ \\
ResNet34 channels-last + compile
& 7.237 ms/batch
& 5.608 ms/batch
& 1.290$\times$ \\
ViT-S FP32
& 43.999 ms/batch
& 37.404 ms/batch
& 1.176$\times$ \\
ViT-S AMP
& 13.728 ms/batch
& 11.860 ms/batch
& 1.157$\times$ \\
ResNet101 FP32
& 16.646 ms/batch
& 14.868 ms/batch
& 1.120$\times$ \\
\bottomrule
\end{tabular}
\end{table*}

As shown in Table~\ref{tab:bp_vs_repl_deploy}, the main inference
benefit comes from the reduced RepL architecture rather than from
operator folding alone. RepL Deploy consistently outperforms the full
BP baseline across CNN and ViT backbones. For ResNet34, the standard
FP32 setting gives a 1.264$\times$ speedup. When using channels-last,
both BP and RepL benefit from a more convolution-friendly memory layout,
and RepL still achieves a 1.273$\times$ speedup. When channels-last is
combined with compilation, both models are executed with a more optimized
static graph, and RepL reaches a 1.290$\times$ speedup. This confirms
that RepL remains beneficial even when strong inference backend
optimizations are applied to the baseline.

For ViT-S, RepL Deploy achieves 1.176$\times$ speedup in FP32 and
1.157$\times$ speedup under AMP. These results indicate that RepL does
not introduce a hidden inference penalty: after deploy
re-parameterization, the replacement layers become ordinary static
operators, and the physically removed blocks translate into practical
inference acceleration.

\paragraph{Fine-tuning on ViTs}
To verify whether RepL remains effective in a transfer-learning setting,
we fine-tune ImageNet-1K~\cite{r4} pretrained ViT-S/16 models on
CIFAR-10~\cite{r1}, SVHN~\cite{r2}, and STL-10~\cite{r3}. The
fine-tuning setting is fixed across datasets: batch size 512, learning
rate $2\times 10^{-4}$, AdamW optimizer, and 100 epochs. Results are
summarized in Table~\ref{tab:finetune_vit}.

\begin{table}[htbp]
\caption{Finetune results on ViT-S/16.}
\label{tab:finetune_vit}
\centering
\scalebox{1.0}{
\setlength{\tabcolsep}{2.0mm}
\begin{tabular}{lllccc}
\toprule
\textbf{Datasets}
& \textbf{Model}
& \textbf{Method}
& \textbf{Acc@1}
& \textbf{\makecell{GPU\\Memory\\(GB)}}
& \textbf{\makecell{Time\\(s/epoch)}} \\
\midrule
\multirow{2}{*}{CIFAR-10}
& \multirow{2}{*}{ViT-S/16}
& E2E
& 95.66 & 25.56 & 32.45 \\
& & \cellcolor[HTML]{E6F0FF}RepL
& \cellcolor[HTML]{E6F0FF}95.89
& \cellcolor[HTML]{E6F0FF}20.14
& \cellcolor[HTML]{E6F0FF}25.18 \\
\midrule
\multirow{2}{*}{SVHN}
& \multirow{2}{*}{ViT-S/16}
& E2E
& 96.92 & 25.56 & 48.44 \\
& & \cellcolor[HTML]{E6F0FF}RepL
& \cellcolor[HTML]{E6F0FF}96.97
& \cellcolor[HTML]{E6F0FF}20.14
& \cellcolor[HTML]{E6F0FF}38.01 \\
\midrule
\multirow{2}{*}{STL-10}
& \multirow{2}{*}{ViT-S/16}
& E2E
& 94.88 & 25.56 & 5.91 \\
& & \cellcolor[HTML]{E6F0FF}RepL
& \cellcolor[HTML]{E6F0FF}95.11
& \cellcolor[HTML]{E6F0FF}20.14
& \cellcolor[HTML]{E6F0FF}4.66 \\
\bottomrule
\end{tabular}}
\end{table}

Table~\ref{tab:finetune_vit} shows that RepL consistently improves or
matches fine-tuning accuracy while reducing training resources. On
CIFAR-10 and STL-10, the accuracy gain is 0.23 percentage points, while
SVHN improves slightly from 96.92 to 96.97. More importantly, memory
usage is reduced from 25.56 GB to 20.14 GB across all three datasets,
and per-epoch time is reduced by around 21--22\%. These results indicate
that the replacement layers do not destroy transferable representations
learned during ImageNet pretraining. Instead, adjacent-parameter synthesis preserves the backbone structure sufficiently well for
downstream adaptation.

\paragraph{Fine-tuning for downstream tasks}
We further evaluate downstream dense prediction by fine-tuning
ImageNet-1K~\cite{r4} pretrained RepL models on CityScapes. We use the
SGD optimizer with batch size 16, learning rate 0.1, crop size 768, and
train for 30k iterations on a single GPU. The results are shown in
Table~\ref{tab:cityscapes_finetune}.

\begin{table*}[t]
\caption{Performance comparison on CityScapes using different backbones.}
\label{tab:cityscapes_finetune}
\centering
\scalebox{1}{
\setlength{\tabcolsep}{2.0mm}
\begin{tabular}{ccccccc}
\toprule
\textbf{Backbone}
& \textbf{Method}
& \textbf{\makecell{Overall\\Accuracy}}
& \textbf{\makecell{Mean\\Accuracy}}
& \textbf{\makecell{Mean\\IoU}}
& \textbf{\makecell{GPU\\Memory\\(GB)}}
& \textbf{\makecell{Time\\(s/epoch)}} \\
\midrule
\multirow{2}{*}{DeepLabV3-R50}
& E2E
& 95.27 & 80.83 & 73.34 & 23.90 & 80 \\
& \cellcolor[HTML]{E6F0FF}RepL
& \cellcolor[HTML]{E6F0FF}95.32
& \cellcolor[HTML]{E6F0FF}81.14
& \cellcolor[HTML]{E6F0FF}73.81
& \cellcolor[HTML]{E6F0FF}20.28
& \cellcolor[HTML]{E6F0FF}68 \\
\midrule
\multirow{2}{*}{DeepLabV3Plus-R50}
& E2E
& 95.66 & 81.89 & 74.61 & 26.81 & 82 \\
& \cellcolor[HTML]{E6F0FF}RepL
& \cellcolor[HTML]{E6F0FF}95.71
& \cellcolor[HTML]{E6F0FF}82.21
& \cellcolor[HTML]{E6F0FF}75.25
& \cellcolor[HTML]{E6F0FF}22.67
& \cellcolor[HTML]{E6F0FF}69 \\
\midrule
\multirow{2}{*}{DeepLabV3-R101}
& E2E
& 95.51 & 82.31 & 74.41 & 30.91 & 95 \\
& \cellcolor[HTML]{E6F0FF}RepL
& \cellcolor[HTML]{E6F0FF}95.54
& \cellcolor[HTML]{E6F0FF}82.71
& \cellcolor[HTML]{E6F0FF}74.55
& \cellcolor[HTML]{E6F0FF}25.90
& \cellcolor[HTML]{E6F0FF}82 \\
\midrule
\multirow{2}{*}{DeepLabV3Plus-R101}
& E2E
& 95.84 & 83.24 & 75.53 & 34.42 & 101 \\
& \cellcolor[HTML]{E6F0FF}RepL
& \cellcolor[HTML]{E6F0FF}95.89
& \cellcolor[HTML]{E6F0FF}84.02
& \cellcolor[HTML]{E6F0FF}76.31
& \cellcolor[HTML]{E6F0FF}28.92
& \cellcolor[HTML]{E6F0FF}86 \\
\bottomrule
\end{tabular}}
\end{table*}

Table~\ref{tab:cityscapes_finetune} shows that RepL improves mean IoU
on all four segmentation backbones while reducing memory and training
time. For example, DeepLabV3Plus-R101 improves the mean IoU from 75.53 to
76.31, while memory decreases from 34.42 GB to 28.92 GB and runtime
decreases from 101 to 86 s/epoch. The improvements in overall accuracy
are relatively small, which is expected because pixel accuracy is
already high on CityScapes. However, the consistent gains in mean
accuracy and mean IoU indicate that RepL preserves spatially useful
representations. This supports the claim that RepL is applicable not
only to classification but also to dense prediction tasks requiring
fine-grained spatial features.

\paragraph{Adaptive replacement strategy}
To test whether learning where to replace blocks can improve the
accuracy--efficiency trade-off, we implement an adaptive variant of
RepL. In this variant, computing layers are attached to internal blocks,
and a binary mask $\mathbf{m}\in\{0,1\}^{n}$ selects which blocks are
replaced in each epoch. For a fair comparison with Fixed-RepL, the number of replaced blocks is kept the same:
\begin{equation}
  \sum_{i=1}^{n} m_i
  =
  |\mathcal{F}|,
\end{equation}
where $|\mathcal{F}|$ is the number of replaced sites in Fixed-RepL.
During training, we compute a per-block redundancy/similarity score and
select the top-$|\mathcal{F}|$ internal blocks. Table~\ref{tab:adaptive_repl}
compares Fixed-RepL and Adaptive-RepL on CIFAR-10.

\begin{table}[t]
\centering
\caption{Comparison between fixed and adaptive replacement strategies on CIFAR-10.}
\renewcommand{\arraystretch}{1.2}
\label{tab:adaptive_repl}
\scalebox{0.96}{
\setlength{\tabcolsep}{1.1mm}
\begin{tabular}{lllccc}
\toprule
\textbf{Data}
& \textbf{Backbone}
& \textbf{Method}
& \textbf{ACC@1}
& \textbf{\makecell{GPU\\Memory\\(GB)}}
& \textbf{\makecell{Training\\Time\\(s/epoch)}} \\
\midrule
\cellcolor[HTML]{E6F0FF}CIFAR-10
& \cellcolor[HTML]{E6F0FF}ResNet-32
& \cellcolor[HTML]{E6F0FF}Fixed-RepL (Ours)
& \cellcolor[HTML]{E6F0FF}93.43
& \cellcolor[HTML]{E6F0FF}2.69
& \cellcolor[HTML]{E6F0FF}8.20 \\
CIFAR-10
& ResNet-32
& Adaptive-RepL
& 92.54 & 3.17 & 9.12 \\
\cellcolor[HTML]{E6F0FF}CIFAR-10
& \cellcolor[HTML]{E6F0FF}ViT-Tiny/8
& \cellcolor[HTML]{E6F0FF}Fixed-RepL (Ours)
& \cellcolor[HTML]{E6F0FF}73.71
& \cellcolor[HTML]{E6F0FF}2.08
& \cellcolor[HTML]{E6F0FF}5.65 \\
CIFAR-10
& ViT-Tiny/8
& Adaptive-RepL
& 71.97 & 2.35 & 6.17 \\
\bottomrule
\end{tabular}}
\end{table}

As shown in Table~\ref{tab:adaptive_repl}, Adaptive-RepL is consistently
worse than Fixed-RepL on both backbones. For ResNet-32, accuracy
decreases by 0.89 points, memory increases by 0.48 GB, and runtime
increases by 0.92 s/epoch. For ViT-Tiny/8, accuracy decreases by 1.74
points, memory increases by 0.27 GB, and runtime increases by 0.52
s/epoch. This suggests that learning the replacement locations
introduces more overhead than benefit. The likely reason is that dynamic
replacement makes the effective network topology change across epochs,
which weakens stable layer-wise co-adaptation. In addition, early-stage redundancy scores can be noisy, so important blocks may be replaced
before the representation becomes stable. These results support the use
of a simple fixed periodic replacement rule.

\paragraph{Extra ablation study on ViT}
In ViT experiments, RepL uses two learnable coefficients, $\alpha$ and
$\beta$, to combine the preceding and succeeding projection weights. To
test whether two-sided weighting is necessary, we compare it with a
single-parameter variant on ViT-T/8 using CIFAR-10. The results are
shown in Table~\ref{tab:repl_param_ablation}.

\begin{table}[htbp]
\caption{Ablation on the number of parameters in RepL. We use ViT-T/8 on the CIFAR-10 dataset.}
\label{tab:repl_param_ablation}
\centering
\setlength{\tabcolsep}{10pt}
\scalebox{0.96}{
\setlength{\tabcolsep}{3.1mm}
\begin{tabular}{lccc}
\toprule
\textbf{Method}
& \textbf{\makecell{Accuracy\\(\%)}}
& \textbf{\makecell{GPU\\Memory\\(GB)}}
& \textbf{\makecell{Training Time\\(s/epoch)}} \\
\midrule
\cellcolor[HTML]{E6F0FF}RepL(2 parameter)
& \cellcolor[HTML]{E6F0FF}$73.71 \pm 1.08$
& \cellcolor[HTML]{E6F0FF}2.08
& \cellcolor[HTML]{E6F0FF}5.65 \\
RepL(1 parameter)
& $73.09 \pm 0.85$
& 2.08
& 5.65 \\
\bottomrule
\end{tabular}}
\end{table}

Table~\ref{tab:repl_param_ablation} shows that using two coefficients
improves accuracy from $73.09\pm0.85$ to $73.71\pm1.08$, while GPU
memory and training time remain unchanged. This indicates that the gain
does not come from extra computation, but from the ability to weight the
preceding and succeeding layers separately. A single coefficient forces
the two neighbors to contribute in a more restricted way, which weakens
the flexibility of the replacement operator. Therefore, separate
$\alpha$ and $\beta$ coefficients provide a small but useful increase in
expressivity at almost zero resource cost.

\paragraph{RepL is orthogonal to deployment-time quantization}
\label{sec:repl_quantization}
To verify that RepL is compatible with deployment-time inference
optimization, we conduct post-hoc INT8 quantization on ImageNet-1K for
ViT-S/16~\cite{r14} and ResNet-34~\cite{r5}. For ViTs, we apply dynamic
INT8 quantization to linear layers
(\texttt{int8\_dynamic\_linear}); for ResNets, we use FX graph-mode
static INT8 quantization with calibration
(\texttt{int8\_fx\_static}). Table~\ref{tab:repl_quant_imagenet}
reports Top-1/Top-5 accuracy and throughput.

\begin{table*}[htbp]
\centering
\small
\setlength{\tabcolsep}{11pt}
\renewcommand{\arraystretch}{1.05}
\caption{\textbf{Post-hoc INT8 quantization on ImageNet-1K.}
ViT~\cite{r14} uses dynamic INT8 quantization on linear layers
(\texttt{int8\_dynamic\_linear}); ResNet~\cite{r5} uses FX graph-mode
static INT8 quantization with calibration (\texttt{int8\_fx\_static}).
$\Delta$ denotes INT8 minus FP32. Throughput is measured in img/s and
latency in ms/img.}
\label{tab:repl_quant_imagenet}
\begin{tabular}{l c cc cc cc cc}
\toprule
Backbone
& Method
& \multicolumn{2}{c}{FP32 Acc. (\%)}
& \multicolumn{2}{c}{INT8 Acc. (\%)}
& $\Delta$Top-1
& $\Delta$Top-5
& FP32
& INT8 \\
\cmidrule(lr){3-4}
\cmidrule(lr){5-6}
& & Top-1 & Top-5 & Top-1 & Top-5 & (\%) & (\%) & (img/s) & (img/s) \\
\midrule
ViT-S/16
& BP
& 63.760 & 83.864
& 63.242 & 83.290
& -0.518 & -0.574
& 1022.948 & 45.788 \\
\cellcolor[HTML]{E6F0FF}ViT-S/16
& \cellcolor[HTML]{E6F0FF}RepL
& \cellcolor[HTML]{E6F0FF}65.336
& \cellcolor[HTML]{E6F0FF}85.766
& \cellcolor[HTML]{E6F0FF}64.222
& \cellcolor[HTML]{E6F0FF}84.534
& \cellcolor[HTML]{E6F0FF}-1.114
& \cellcolor[HTML]{E6F0FF}-1.232
& \cellcolor[HTML]{E6F0FF}1072.901
& \cellcolor[HTML]{E6F0FF}48.723 \\
ResNet-34
& BP
& 73.660 & 91.152
& 73.496 & 91.086
& -0.164 & -0.066
& 1757.045 & 194.469 \\
\cellcolor[HTML]{E6F0FF}ResNet-34
& \cellcolor[HTML]{E6F0FF}RepL
& \cellcolor[HTML]{E6F0FF}75.112
& \cellcolor[HTML]{E6F0FF}91.914
& \cellcolor[HTML]{E6F0FF}74.832
& \cellcolor[HTML]{E6F0FF}91.710
& \cellcolor[HTML]{E6F0FF}-0.280
& \cellcolor[HTML]{E6F0FF}-0.204
& \cellcolor[HTML]{E6F0FF}1837.255
& \cellcolor[HTML]{E6F0FF}249.975 \\
\bottomrule
\end{tabular}

\vspace{0.35em}
\footnotesize
\textbf{Latency (ms/img).}
ViT-S/16 BP: FP32 0.978, INT8 21.839; \ \ 
ViT-S/16 RepL: FP32 0.932, INT8 20.524.
ResNet-34 BP: FP32 0.569, INT8 5.142; \ \ 
ResNet-34 RepL: FP32 0.544, INT8 4.000.
\end{table*}

Table~\ref{tab:repl_quant_imagenet} shows that RepL remains effective
after quantization. For ViT-S/16, RepL improves Top-1 accuracy over BP
by 1.576 points in FP32 and still keeps a 0.980-point advantage after
INT8 quantization. For ResNet-34, RepL improves Top-1 by 1.452 points
in FP32 and 1.336 points after INT8. The INT8 accuracy drop is also
moderate: RepL drops by 1.114 points on ViT-S/16 and 0.280 points on
ResNet-34. Although the absolute INT8 throughput depends on the backend
and is not the main claim of this experiment, RepL is faster than BP
under the same precision setting in our reference implementation. These
results suggest that RepL can be combined with standard post-training
quantization without losing its accuracy advantage.

\section{Theoretical Analysis}
\label{sec:theory}

This section analyzes why Replacement Learning (RepL) can reduce
training cost while maintaining the optimization behavior of the
original model. We focus on four aspects: parameter count, computation
and memory cost, approximation error, and optimization convergence. We
use $N$ to denote the number of original blocks, $K$ to denote the
replacement interval, and $q$ to denote the spatial convolution kernel
size.

\subsection{Notation and Assumptions}
\label{sec:theory_notation}

Let the original full-depth network be
\begin{equation}
\begin{aligned}
  F(\mathbf{x};\Theta)
  &=
  f_N \circ f_{N-1}
  \circ \cdots \circ f_1(\mathbf{x}),          \\
  \mathbf{h}_{\ell}
  &=
  f_{\ell}
  \!\left(
    \mathbf{h}_{\ell-1};
    \theta_{\ell}
  \right),
  \qquad
  \mathbf{h}_0=\mathbf{x}.
\end{aligned}
\label{eq:theory_full_forward_short}
\end{equation}
RepL physically removes the internal block set
\begin{equation}
\begin{aligned}
  \mathcal{R}
  =
  \left\{
  r
  \ \middle|\
  r=mK,\;
  1\le r<N,\;
  m=1,2,\ldots
  \right\},
\end{aligned}
\label{eq:theory_removed_set_short}
\end{equation}
and inserts a lightweight computing layer $g_r$ at each removed
position. Let
\begin{equation}
\begin{aligned}
  R=|\mathcal{R}|,
  \qquad
  \rho=R/N .
\end{aligned}
\end{equation}
For $K=4$, $\rho\approx 1/4$, up to the boundary effect caused by
keeping the last block.

The RepL network is denoted by
\begin{equation}
\begin{aligned}
  F_{\mathrm{RepL}}
  \!\left(
    \mathbf{x};
    \Theta_{\mathrm{keep}},
    \Psi
  \right),
\end{aligned}
\end{equation}
where $\Theta_{\mathrm{keep}}$ contains retained backbone parameters
and $\Psi=\{\psi_r:r\in\mathcal{R}\}$ contains the additional
computing-layer parameters.

We use the following local assumptions along the training trajectory.

\paragraph{A1: compatible replacement sites.}
Each removed block has compatible left and right neighbors. For CNNs,
replacement is performed inside the same residual stage. For ViTs,
token length and embedding dimension are unchanged across adjacent
blocks.

\paragraph{A2: bounded activations and stable suffixes.}
For each $r\in\mathcal{R}$,
\begin{equation}
\begin{aligned}
  \left\|
    \mathbf{h}_{r-1}
  \right\|
  \le H_r .
\end{aligned}
\label{eq:bounded_activation_short}
\end{equation}
The suffix after position $r$ has local Lipschitz factor
\begin{equation}
\begin{aligned}
  \Pi_r
  =
  \prod_{\ell>r}
  L_{\ell}^{\star},
\end{aligned}
\label{eq:suffix_lipschitz_short}
\end{equation}
where $L_{\ell}^{\star}$ is the local Lipschitz constant of block
$\ell$ on the feature domain visited during training.

\paragraph{A3: local replacement error.}
For each removed block $r$, define
\begin{equation}
\begin{aligned}
  \varepsilon_r
  =
  \sup_{\mathbf{h}\in\mathcal{D}_r}
  \frac{
    \left\|
      g_r(\mathbf{h})
      -
      f_r(\mathbf{h})
    \right\|
  }{
    \max
    \left\{
      \left\|\mathbf{h}\right\|,
      1
    \right\}
  },
\end{aligned}
\label{eq:local_replacement_error_short}
\end{equation}
where $\mathcal{D}_r$ is the activation domain at position $r$. This
block-level error includes normalization, nonlinearities, and residual
additions.

\subsection{Parameter, Computation, and Memory Reduction}
\label{sec:theory_cost_short}

Let $P_{\ell}$ be the parameter count of the original block
$f_{\ell}$, and let $a_r$ be the additional parameter count introduced
by the computing layer at position $r$. Since RepL physically removes
the blocks in $\mathcal{R}$, its parameter count is
\begin{equation}
\begin{aligned}
  P_{\mathrm{RepL}}
  &=
  \sum_{\ell\notin\mathcal{R}}
  P_{\ell}
  +
  \sum_{r\in\mathcal{R}}
  a_r                                                   \\
  &=
  P_{\mathrm{E2E}}
  -
  \sum_{r\in\mathcal{R}}
  \left(
    P_r-a_r
  \right).
\end{aligned}
\label{eq:repl_param_exact_short}
\end{equation}
Therefore, RepL reduces trainable parameters whenever
$a_r<P_r$, which holds for standard CNN and ViT blocks.

\paragraph{CNN BasicBlock.}
For a same-stage BasicBlock with channel width $C_r$, the original
block contains two $q\times q$ convolutions and two affine BN layers:
\begin{equation}
\begin{aligned}
  P_r^{\mathrm{basic}}
  =
  2C_r^2q^2
  +
  4C_r .
\end{aligned}
\end{equation}
The computing layer uses channel-wise coefficients
$\alpha_{r,c},\beta_{r,c}$ and one affine BN layer:
\begin{equation}
\begin{aligned}
  a_r^{\mathrm{basic}}
  =
  2C_r + 2C_r
  =
  4C_r .
\end{aligned}
\end{equation}
Thus,
\begin{equation}
\begin{aligned}
  \frac{
    a_r^{\mathrm{basic}}
  }{
    P_r^{\mathrm{basic}}
  }
  \le
  \frac{
    2
  }{
    C_r q^2
  },
\end{aligned}
\label{eq:basic_param_ratio_short}
\end{equation}
which is small for normal channel widths and $q=3$.

\paragraph{CNN Bottleneck.}
For a Bottleneck with input/output width $C_r$ and middle width $B_r$,
\begin{equation}
\begin{aligned}
  P_r^{\mathrm{bottle}}
  =
  2C_rB_r
  +
  B_r^2q^2
  +
  4B_r
  +
  2C_r .
\end{aligned}
\end{equation}
RepL reuses adjacent projection weights as synthesis anchors and only
learns lightweight synthesis coefficients and one BN:
\begin{equation}
\begin{aligned}
  a_r^{\mathrm{bottle}}
  =
  2B_r
  +
  2C_r .
\end{aligned}
\end{equation}
Hence,
\begin{equation}
\begin{aligned}
  P_r^{\mathrm{bottle}}
  -
  a_r^{\mathrm{bottle}}
  =
  2C_rB_r
  +
  B_r^2q^2
  +
  2B_r
  >
  0 .
\end{aligned}
\label{eq:bottle_param_saving_short}
\end{equation}

\paragraph{ViT block.}
For a ViT block with embedding dimension $d$, the dominant parameter
count is
\begin{equation}
\begin{aligned}
  P_r^{\mathrm{ViT}}
  =
  12d^2
  +
  \mathcal{O}(d),
\end{aligned}
\end{equation}
from Q/K/V projections, output projection, and the MLP. The basic ViT computing layer introduces two learnable synthesis coefficients:
\begin{equation}
\begin{aligned}
  a_r^{\mathrm{ViT}}=2,
\end{aligned}
\end{equation}
or $a_r^{\mathrm{ViT}}=2H$ if head-wise synthesis with $H$ heads is
used. Thus,
\begin{equation}
\begin{aligned}
  \frac{
    a_r^{\mathrm{ViT}}
  }{
    P_r^{\mathrm{ViT}}
  }
  =
  \mathcal{O}
  \!\left(
    \frac{H}{d^2}
  \right),
\end{aligned}
\end{equation}
which is negligible for common ViT widths.

For nearly homogeneous blocks, $P_{\ell}\approx\overline{P}$ and
$R/N\approx 1/K$, so
\begin{equation}
\begin{aligned}
  \frac{
    P_{\mathrm{RepL}}
  }{
    P_{\mathrm{E2E}}
  }
  \approx
  1
  -
  \frac{1}{K}
  \left(
    1
    -
    \frac{
      \overline{a}_{\mathcal{R}}
    }{
      \overline{P}
    }
  \right).
\end{aligned}
\label{eq:param_homogeneous_short}
\end{equation}
Since $\overline{a}_{\mathcal{R}}\ll\overline{P}$, the parameter reduction is close to the physical removal ratio.

Let $F_{\ell}$ be the forward FLOPs of the original block, and
$\widetilde{F}_r$ be the forward FLOPs of the computing layer. The block-level FLOP relation is
\begin{equation}
\begin{aligned}
  F_{\mathrm{RepL}}
  =
  F_{\mathrm{E2E}}
  -
  \sum_{r\in\mathcal{R}}
  F_r
  +
  \sum_{r\in\mathcal{R}}
  \widetilde{F}_r
  +
  F_{\mathrm{synth}},
\end{aligned}
\label{eq:flop_exact_short}
\end{equation}
where $F_{\mathrm{synth}}$ is the weight-synthesis cost. This term is
independent of image resolution or token length, and is usually small.

Define
\begin{equation}
\begin{aligned}
  \eta_r
  =
  \frac{
    \widetilde{F}_r
  }{
    F_r
  } .
\end{aligned}
\end{equation}
For homogeneous blocks,
\begin{equation}
\begin{aligned}
  \frac{
    F_{\mathrm{RepL}}
  }{
    F_{\mathrm{E2E}}
  }
  \approx
  1
  -
  \frac{1}{K}
  \left(
    1-\overline{\eta}
  \right).
\end{aligned}
\label{eq:flop_homogeneous_short}
\end{equation}

For a CNN BasicBlock, two $q\times q$ convolutions are replaced by one
synthesized $q\times q$ convolution, giving
\begin{equation}
\begin{aligned}
  \eta_r^{\mathrm{basic}}
  \approx
  \frac{1}{2}.
\end{aligned}
\end{equation}
For a basic ViT computing layer, a full attention--MLP block is replaced
by one token-wise projection:
\begin{equation}
\begin{aligned}
  \eta_r^{\mathrm{ViT}}
  &\approx
  \frac{
    2Td^2
  }{
    24Td^2+4T^2d
  }                                      \\
  &=
  \frac{
    1
  }{
    12+2T/d
  } ,
\end{aligned}
\label{eq:vit_eta_short}
\end{equation}
where $T$ is the token number. Therefore, ViT replacement removes the
dominant self-attention and MLP computation of the original block.

Training memory is dominated by stored activations. Let $S_r$ and
$\widetilde{S}_r$ be the activation memory of the original block and
the computing layer. Then
\begin{equation}
\begin{aligned}
  M_{\mathrm{act}}^{\mathrm{RepL}}
  \le
  M_{\mathrm{act}}^{\mathrm{E2E}}
  -
  \sum_{r\in\mathcal{R}}
  \left(
    S_r-\widetilde{S}_r
  \right)
  +
  M_{\mathrm{aux}},
\end{aligned}
\label{eq:memory_general_short}
\end{equation}
where $M_{\mathrm{aux}}$ contains small auxiliary states such as
synthesis coefficients and normalization statistics. In ViTs, this is
especially beneficial because the removed attention matrix has memory
order $\mathcal{O}(BHT^2)$, while the basic computing layer only stores
token-wise activations of order $\mathcal{O}(BTd)$.

\subsection{Approximation Error and Gradient Bias}
\label{sec:theory_error_short}

We next bound how the local replacement errors accumulate. Under
Assumptions A1--A3, for any input $\mathbf{x}$,
\begin{equation}
\begin{aligned}
&
  \left\|
    F_{\mathrm{RepL}}(\mathbf{x})
    -
    F(\mathbf{x})
  \right\|
\\
&\quad
  \le
  \sum_{r\in\mathcal{R}}
  \Pi_r
  \varepsilon_r
  \max
  \left\{
    H_r,
    1
  \right\}.
\end{aligned}
\label{eq:output_deviation_short}
\end{equation}
This is followed by replacing the removed blocks one by one: the local
error created at position $r$ is bounded by
$\varepsilon_r\max\{H_r,1\}$ and is amplified by the suffix factor
$\Pi_r$.

If the suffixes are locally non-expansive, i.e., $\Pi_r\le1$, and
$\varepsilon_r\le\varepsilon_{\max}$, $H_r\le H_{\max}$, then
\begin{equation}
\begin{aligned}
  \left\|
    F_{\mathrm{RepL}}(\mathbf{x})
    -
    F(\mathbf{x})
  \right\|
  \le
  R
  \varepsilon_{\max}
  \max
  \left\{
    H_{\max},
    1
  \right\}.
\end{aligned}
\label{eq:output_linear_short}
\end{equation}
Since $R\approx N/K$, the accumulated discrepancy is controlled by the
replacement interval.

Let
\begin{equation}
\begin{aligned}
  \ell(\Theta)
  &=
  \mathbb{E}
  \left[
    \mathcal{L}
    \!\left(
      F(\mathbf{x};\Theta),
      y
    \right)
  \right],                                      \\
  J(\Theta_{\mathrm{keep}},\Psi)
  &=
  \mathbb{E}
  \left[
    \mathcal{L}
    \!\left(
      F_{\mathrm{RepL}}
      \!\left(
        \mathbf{x};
        \Theta_{\mathrm{keep}},
        \Psi
      \right),
      y
    \right)
  \right].
\end{aligned}
\label{eq:losses_short}
\end{equation}
Let $\Theta_s$ denote the shared retained parameters. We assume the
gradient map with respect to $\Theta_s$ is locally Lipschitz in the
network output:
\begin{equation}
\begin{aligned}
&
  \left\|
    \nabla_{\Theta_s}
    \mathcal{L}(\mathbf{z},y)
    -
    \nabla_{\Theta_s}
    \mathcal{L}(\mathbf{z}',y)
  \right\|
\\
&\quad
  \le
  G_{\Theta}
  \left\|
    \mathbf{z}
    -
    \mathbf{z}'
  \right\|.
\end{aligned}
\label{eq:grad_lipschitz_short}
\end{equation}
Combining this condition with Eq.~\eqref{eq:output_deviation_short}
gives
\begin{equation}
\begin{aligned}
  \left\|
    \nabla_{\Theta_s}J
    -
    \nabla_{\Theta_s}\ell
  \right\|
  \le
  B_{\mathrm{grad}},
\end{aligned}
\label{eq:bgrad_short_short}
\end{equation}
where
\begin{equation}
\begin{aligned}
  B_{\mathrm{grad}}
  =
  G_{\Theta}
  \,
  \mathbb{E}_{\mathbf{x}}
  \left[
    \sum_{r\in\mathcal{R}}
    \Pi_r
    \varepsilon_r
    \max
    \left\{
      H_r,
      1
    \right\}
  \right].
\end{aligned}
\label{eq:bgrad_def_short}
\end{equation}
Thus, the gradient bias introduced by RepL is controlled by local
replacement errors and suffix stability.

\subsection{Optimization Convergence}
\label{sec:theory_convergence_short}

RepL is optimized end-to-end on its own objective $J$. Assume $J$ is
$L_J$-smooth and that stochastic gradients are unbiased with variance
bounded by $\sigma^2/B$:
\begin{equation}
\begin{aligned}
  \mathbb{E}[\mathbf{g}_t]
  &=
  \nabla J(\Theta_t,\Psi_t),                         \\
  \mathbb{E}
  \left[
    \left\|
      \mathbf{g}_t
      -
      \nabla J(\Theta_t,\Psi_t)
    \right\|^2
  \right]
  &\le
  \frac{\sigma^2}{B}.
\end{aligned}
\label{eq:sgd_assumption_short}
\end{equation}
Running SGD with $\eta\le 1/L_J$ for $T$ iterations yields the standard
nonconvex bound
\begin{equation}
\begin{aligned}
  \frac{1}{T}
  \sum_{t=0}^{T-1}
  \mathbb{E}
  \left[
    \left\|
      \nabla J(\Theta_t,\Psi_t)
    \right\|^2
  \right]
  \le
  \frac{
    2(J_0-J^{\star})
  }{
    \eta T
  }
  +
  \frac{
    \eta L_J\sigma^2
  }{
    B
  } .
\end{aligned}
\label{eq:sgd_repl_convergence_short}
\end{equation}
With $\eta=\mathcal{O}(\sqrt{B/T})$, this gives
\begin{equation}
\begin{aligned}
  \frac{1}{T}
  \sum_{t=0}^{T-1}
  \mathbb{E}
  \left[
    \left\|
      \nabla J(\Theta_t,\Psi_t)
    \right\|^2
  \right]
  =
  \mathcal{O}
  \!\left(
    \frac{1}{\sqrt{BT}}
  \right).
\end{aligned}
\label{eq:sgd_rate_short}
\end{equation}

The stationary behavior with respect to the original full network objective can be related through the gradient-bias term. From
Eq.~\eqref{eq:bgrad_short_short},
\begin{equation}
\begin{aligned}
&
\min_{0\le t<T}
\mathbb{E}
\Bigl[
  \bigl\|
    \nabla_{\Theta_s}
    \ell(\Theta_t)
  \bigr\|^2
\Bigr]
\\
&\quad\le
2
\min_{0\le t<T}
\mathbb{E}
\Bigl[
  \bigl\|
    \nabla_{\Theta_s}
    J(\Theta_t,\Psi_t)
  \bigr\|^2
\Bigr]
+
2B_{\mathrm{grad}}^2 .
\end{aligned}
\label{eq:stationarity_transfer_short}
\end{equation}
Combining Eq.~\eqref{eq:sgd_rate_short} and
Eq.~\eqref{eq:stationarity_transfer_short} gives
\begin{equation}
\begin{aligned}
&
\min_{0\le t<T}
\mathbb{E}
\Bigl[
  \bigl\|
    \nabla_{\Theta_s}
    \ell(\Theta_t)
  \bigr\|^2
\Bigr]
\\
&\quad
=
\mathcal{O}
\!\left(
  \frac{1}{\sqrt{BT}}
\right)
+
\mathcal{O}
\!\left(
  B_{\mathrm{grad}}^2
\right).
\end{aligned}
\label{eq:stationarity_rate_short}
\end{equation}
Therefore, RepL keeps the standard nonconvex convergence behavior for
its own training objective, and its stationary point is close to that
of the original objective when the replacement error is small.

\subsection{Recoverability of the Computing Layer}
\label{sec:theory_recoverability_short}

The previous bounds depend on $\varepsilon_r$. We now relate
$\varepsilon_r$ to the expressivity of the neighbor-synthesized
computing layer.

\paragraph{CNN case.}
For a removed BasicBlock, RepL synthesizes the convolutional kernel by
\begin{equation}
\begin{aligned}
  \widehat{W}_{r,c,:,:}
  =
  \alpha_{r,c}
  \overline{W}_{\mathrm{prev},c,:,:}
  +
  \beta_{r,c}
  \overline{W}_{\mathrm{next},c,:,:}.
\end{aligned}
\label{eq:cnn_recover_short}
\end{equation}
Define the representable kernel set
\begin{equation}
\begin{aligned}
  \mathcal{S}_r^{\mathrm{CNN}}
  =
  \left\{
    \widehat{W}_r
    \ \middle|\
    \widehat{W}_{r,c,:,:}
    =
    \alpha_{r,c}
    \overline{W}_{\mathrm{prev},c,:,:}
    +
    \beta_{r,c}
    \overline{W}_{\mathrm{next},c,:,:}
  \right\}.
\end{aligned}
\label{eq:cnn_span_short}
\end{equation}
The local replacement error is the distance from the removed block to
the function class induced by $\mathcal{S}_r^{\mathrm{CNN}}$:
\begin{equation}
\begin{aligned}
  \varepsilon_r
  =
  \operatorname{dist}_{\mathcal{D}_r}
  \left(
    f_r,
    \mathcal{G}_r^{\mathrm{CNN}}
  \right).
\end{aligned}
\label{eq:cnn_distance_short}
\end{equation}
If $f_r\in\mathcal{G}_r^{\mathrm{CNN}}$ on $\mathcal{D}_r$, then there
exist synthesis coefficients such that $\varepsilon_r=0$.

\paragraph{ViT case.}
For the basic ViT computing layer,
\begin{equation}
\begin{aligned}
  \widehat{W}_{r}^{\mathrm{proj}}
  =
  \alpha_r
  W_{\mathrm{prev}}^{\mathrm{proj}}
  +
  \beta_r
  W_{\mathrm{next}}^{\mathrm{proj}} .
\end{aligned}
\label{eq:vit_recover_short}
\end{equation}
The best projection approximation error is
\begin{equation}
\begin{aligned}
  d_r^{\mathrm{proj}}
  =
  \min_{\alpha,\beta}
  \left\|
    W_r^{\mathrm{proj}}
    -
    \alpha W_{\mathrm{prev}}^{\mathrm{proj}}
    -
    \beta W_{\mathrm{next}}^{\mathrm{proj}}
  \right\|_F .
\end{aligned}
\label{eq:vit_proj_distance_short}
\end{equation}
For token features
$\mathbf{X}\in\mathbb{R}^{B\times T\times d}$,
\begin{equation}
\begin{aligned}
&
  \left\|
    \mathbf{X}
    \left(
      W_r^{\mathrm{proj}}
      -
      \widehat{W}_{r}^{\mathrm{proj}}
    \right)^{\top}
  \right\|_F
\\
&\quad
  \le
  \left\|
    \mathbf{X}
  \right\|_F
  d_r^{\mathrm{proj}} .
\end{aligned}
\label{eq:vit_projection_output_short}
\end{equation}
Thus, the ViT replacement error is controlled by the distance between
the removed projection and the span of its neighboring projections. If the removed projection lies in this span, the token-wise projection error can be zero.

\subsection{Replacement Interval Trade-off}
\label{sec:theory_k_short}

The replacement interval $K$ controls the cost--bias trade-off. From
Eq.~\eqref{eq:flop_homogeneous_short}, the per-epoch training cost can
be approximated as
\begin{equation}
\begin{aligned}
  C(K)
  \approx
  C_0
  \left[
    1
    -
    \frac{
      1-\overline{\eta}
    }{
      K
    }
  \right],
\end{aligned}
\label{eq:cost_k_short}
\end{equation}
where $C_0$ is the full model cost. Meanwhile, from
Eq.~\eqref{eq:bgrad_def_short}, if
$\varepsilon_r\le\overline{\varepsilon}$ and
$\Pi_r\le\Pi_{\max}$, then
\begin{equation}
\begin{aligned}
  B_{\mathrm{grad}}(K)
  \le
  G_{\Theta}
  \Pi_{\max}
  \max
  \left\{
    H_{\max},
    1
  \right\}
  \frac{N}{K}
  \overline{\varepsilon}.
\end{aligned}
\label{eq:bgrad_k_short}
\end{equation}
Thus, a smaller $K$ removes more blocks and reduces cost, while a larger
$K$ reduces the accumulated replacement bias. The default setting
$K=4$ balances these two effects in our experiments.

\subsection{Summary}
\label{sec:theory_summary_short}

The analysis shows that RepL reduces trainable parameters because each
physically removed block is replaced by a much smaller computing layer.
Its computation and memory savings follow from removing full block
operations and their intermediate activations. The approximation and
optimization behavior is controlled by the local errors
$\varepsilon_r$. When neighboring retained blocks provide a good
synthesis basis, these errors are small, and RepL preserves standard
end-to-end optimization behavior while reducing training cost.

\section{Conclusion}
This paper introduces a novel learning approach called Replacement Learning (RepL), designed to address the challenge of maintaining model performance while reducing computational overhead and resource consumption. Replacement Learning effectively reduces the parameter count by removing specific layers and replacing them with computing layers. These computing layers integrate the outputs of the preceding and subsequent layers, enhancing the integration of low-level and high-level features, thereby improving the overall performance of the model. We apply Replacement Learning to various model architectures with different depths and evaluate their performance on five widely used datasets in classification and object detection tasks. Results demonstrate that the proposed RepL not only reduces training time and GPU usage but also consistently outperforms end-to-end training in terms of overall performance.

\bibliographystyle{IEEEtran}
\bibliography{IEEEabrv,ref}

\end{document}